\documentclass[final]{siamltex}

\usepackage[]{fontenc}
\usepackage[latin1]{inputenc}
\usepackage{verbatim}
\usepackage{amsmath}
\usepackage{amssymb}
\usepackage{dsfont}
\usepackage{amsfonts}
\usepackage{mathrsfs}
\usepackage{graphicx}
\usepackage{color}
\usepackage{ulem}
\usepackage{pgfplots}
\usepackage{setspace}
\usepackage{xspace}
\usepackage{eclbkbox}

\usepackage{ stmaryrd }
\usepackage{graphicx}
\usepackage{lineno}
\usepackage{color}
\usepackage{url}
\usepackage{graphicx}
\usepackage{ stmaryrd }

\usepackage{amsmath,amssymb,color} 

\DeclareMathOperator*{\argmin}{arg\,min}

\newcommand\epsi{\boldsymbol{\epsilon}}

\newcommand\xx{\mathbf{x}}
\newcommand\yy{\mathbf{y}}

\newcommand{\dd}{\mathbf{d}}
\newcommand{\cv}{\mathbf{c}}

\newcommand{\zz}{\boldsymbol{z}}
\newcommand{\zv}{\boldsymbol{\zeta}}

\newcommand{\uv}{\mathbf{u}}

\newcommand{\Am}{\mathbf{A}}

\newcommand{\Bm}{\mathbf{B}}

\newcommand{\Cc}{\Omega}

\newcommand{\Rr}{\mathds{R}}
\newcommand{\Zc}{\Xi}

\newcommand{\eg}{\textit{e.g.}}
\newcommand{\ie}{\textit{i.e.}}

\newcommand{\Rb}{\mathds{R}}

\newlength{\textlarg}

\newcommand\transpose[1][{}]{{#1}^{*}}

\title{An Efficient Algorithm for Video Super-Resolution Based On a Sequential Model}

\author{P. H\'EAS\thanks{INRIA Centre Rennes - Bretagne Atlantique, campus universitaire de Beaulieu, 35042 Rennes, France ({\tt patrick.heas@inria.fr, cedric.herzet@inria.fr}) }   \and A. DR\'EMEAU\thanks{ENSTA Bretagne,
2 rue Francois Verny, 29806 Brest, France$\qquad$ ({\tt angelique.dremeau@ensta-bretagne.fr})} \and   C. HERZET$^*$
   }
  
\makeatletter 
\renewcommand{\thefigure}{\@arabic\c@figure}
\makeatother
\begin{document}

\maketitle

\begin{abstract}
{In this work, we propose a novel procedure for video super-resolution, that is the recovery of a sequence of high-resolution images from its low-resolution counterpart. Our approach is based on a ``sequential'' model (\ie, each high-resolution frame is supposed to be a displaced version {of} the preceding one) and considers the use of sparsity-enforcing priors.  Both the recovery of the high-resolution images and the motion fields relating them is tackled. This leads to a large-dimensional, non-convex and non-smooth problem.} 
{We propose an algorithmic framework to address the latter.} 
Our approach relies on fast gradient evaluation methods and modern optimization techniques for non-differentiable/non-convex problems. Unlike some other previous works, we show that there exists a provably-convergent method  with a complexity linear in the problem dimensions. 
We assess the proposed optimization method on {several video benchmarks and emphasize its good performance with respect to the state of the art.}

\end{abstract}


\pagestyle{myheadings}
\thispagestyle{plain}
\markboth{}{An Efficient Algorithm for Video Super-Resolution}

\section{Introduction}
Super resolution (SR) aims at reconstructing  high-resolution (HR) images from distorted low-resolution (LR) observations. 
 This type of methodology dates back to the 70's with the pioneering work of Gerchberg \cite{Gerchberg1974Superresolution} and  Santis\&Gori \cite{DeSantis1975Iterative}. Since then, super resolution has been applied to a large variety of applicative domains, including infrared \cite{Hardie1998Highresolution}, medical \cite{Schatzberg1992Superresolution}, satellite and aerial \cite{Peleg1987Improving,Tsai1984Multiframe} imaging.  We refer the reader to \cite{Nasrollahi2014Superresolution} for a pretty comprehensive overview of the works dealing with SR. 
 
{One can distinguish between different setups in the domain of super resolution. ``Single{-}frame'' super resolution aims at computing an enhanced version {of} some HR image from the observation of \textit{one} single LR image, see \eg,  {\cite{Dong2013Nonlocally, He2013Beta,Peleg14}. On the other hand, the ``multi-frame'' paradigm typically focusses on the recovery of \textit{one} HR image by exploiting the observations of \textit{several} LR frames, see \eg, \cite{Elad1995Superresolution,Patti1997Superresolution,Farsiu2004Fast,Fransens2007Optical,Mitzel09, Unger2010Convex,Liu2014Bayesian}. Finally, the ``video''  super resolution problem consists in estimating a \textit{sequence} of HR images from the observations of their LR counterparts. We consider the latter paradigm in this paper. }

From a conceptual point of view, a simple (but valid) solution to address video super-resolution consists in applying single-frame or multi-frame procedures on each frame of the HR sequence to recover. 
 This strategy was for example considered in \cite{Elad1995Superresolution,Patti1997Superresolution,Farsiu2004Fast,Fransens2007Optical,Mitzel09, Unger2010Convex,Liu2014Bayesian}. 
  Nevertheless, this approach may fail in properly exploiting the strong temporal correlations existing between the (successive) frames of the HR sequence. 
  Hence, procedures specifically dedicated to accounting for these dependencies have been proposed in the literature, see \cite{Tom1994Reconstruction,Elad1999Superresolution,Elad1999SuperresolutionB,Wang2002Superresolution,Zhao2002SuperResolution,Farsiu2006VideotoVideo,Hardie2007Fast, Takeda09,Milanfar10}.  
  A central element   is the ``sequential'' model linking the frames of the HR sequence. 
  More specifically, in  most of these methods, the frames of the HR sequence are supposed to obey a dynamical model where each HR image is seen as a displaced version (by some unknown motion field) of the preceding one (see section \ref{sec:2} for a detailed description). 
  This is in contrast with the standard multi-frame model where each LR observation is assumed to be a LR displaced version of one given reference frame. 

The practical exploitation of the ``sequential'' model nevertheless faces a certain number of bottlenecks. The most stringent one is probably the model  dimensionality: because it accounts for the temporal evolution of each HR frame, the number of variables involved in the sequential model may become very large. This makes video SR based on sequential models pretty challenging. As a matter of fact, in comparison with the huge number of papers dealing with SR, only a few have focussed on  this particular problem,  see \cite{Tom1994Reconstruction,Elad1999Superresolution,Elad1999SuperresolutionB,Wang2002Superresolution,Zhao2002SuperResolution,Farsiu2006VideotoVideo,Hardie2007Fast, Takeda09,Milanfar10}. 
 In \cite{Tom1994Reconstruction}, the authors modeled the dependence between the different images of the sequence as a Gaussian process and provided an efficient implementation in the Fourier domain. Other contributions  relied on adaptive-filtering techniques, see \cite{Elad1999Superresolution,Elad1999SuperresolutionB,Wang2002Superresolution,Zhao2002SuperResolution,Farsiu2006VideotoVideo,Hardie2007Fast}. In this line of thought, most of the contributions cited above considered that the HR sequence is ruled by a state-space sequential model and the authors derived estimation  procedures inspired by the well-known Kalman filter. The standard Kalman updates  leading to a prohibitive complexity in the context of video SR, Elad and co-authors published a series of papers \cite{Elad1999Superresolution,Elad1999SuperresolutionB, Farsiu2006VideotoVideo}  in which they proposed updates having a linear complexity in the  problem dimensions. Their approach is based on some approximations of the model and/or Kalman updates (\eg, uniform translational motion \cite{Farsiu2006VideotoVideo}, noise-free evolution model \cite{Elad1999SuperresolutionB}, etc). In \cite{Takeda09,Milanfar10}, the authors considered a local approximation of the state-space sequential model by using steering kernel regression on the LR observations.

{In this paper, we provide an approximation-free methodological framework exploiting  a sequential model for video SR. We express the unknown HR sequence as the solution of a constrained optimization problem and propose an iterative procedure to solve the latter. Our method is provably convergent (to a local minimum of the problem)  and has a tractable complexity per iteration (\ie,  linear in the problem dimensions).} 
 {The proposed framework encompasses two important ingredients of video SR, namely: \textit{i)} a precise characterization of the motion {fields} linking the successive frames of the sequence; \textit{ii)} the exploitation of proper priors on the unknowns of the problem. These two ingredients lead to additional difficulties (in top of the large dimensionality) since they typically introduce \textit{non-convex} and \textit{non-smooth} terms in the cost function to minimize. We elaborate on these points in the  two next paragraphs.}

 {A precise characterization of the model connecting the different images of the HR sequence is crucial for the success of video SR.} 
 Typically, videos are  characterized  by non-global motions. This is in contrast with many standard SR models of the literature which assume global motions (\eg,  translation \cite{Akgun2005Superresolution}, affine \cite{Yang2009New} or projection \cite{Capel2001Superresolution}), well-suited to still image reconstruction. 
 The imaging model in video SR  thus takes  a more-involved form and has to be considered with care. In particular, the estimation of the motion between two consecutive frames is usually tantamount to solving an optical-flow problem \cite{Baker1999Superresolution}. 
 Embedding motion estimation in the SR reconstruction introduces  new difficulties: \textit{i)} it increases the problem dimensionality since two additional unknowns (the displacement in each direction) have to  be estimated for each pixel of the HR images;  \textit{ii)} it typically introduces non-linearities in the image formation model. 
 These obstacles  are particularly prominent  in the case of a sequential model because of the nested structure of the unknowns  dependencies.
  As a consequence, {until recently,}  motion estimation has been overlooked and considered as a side problem in many SR contributions  involving  either muti-frame or sequential models  (see \eg,  \cite{Patti1997Superresolution,Elad1995Superresolution,Elad1999Superresolution,Elad1999SuperresolutionB,Wang2002Superresolution,Zhao2002SuperResolution,Farsiu2004Fast,Farsiu2006VideotoVideo,Hardie2007Fast,Unger2010Convex,Mitzel09}) with the exception of \eg, \cite{Baker1999Superresolution,Fransens2007Optical,Keller2011Video,Liu14} . 

 Interestingly, several authors have emphasized the importance of accurate motion estimation in the video SR process and provided studies of the sensibility of adaptive filtering techniques to the latter, see \cite{Zhao2002SuperResolution,Costa2006Design,Costa2007Statistical,Costa2009Registration}. 
 In this paper, we show that the motion {estimation} can be included {in our video SR problem} without significantly  increasing the computational cost. 
 
 {Another important ingredient for the success of video SR is the definition of proper priors on (some of) the unknowns of the problem. 
  Indeed, video SR is a naturally ill-posed problem: typical setups impose the observation of (at most) one LR image per frame of the HR sequence; 
  hence, if the motion between the different frames is unknown, it is easy to see that the number of variables which have to be  estimated is well beyond the number of observations.} 
  In order to tackle this difficulty, a well-known technique  consists in resorting to prior information on the sought quantities. 
  This type of approach has been used extensively (but not only) in the context of single-image SR, where an HR image has to be reconstructed from one single LR observation. 
  First methodologies based on prior information date back to the 70's \cite{Gerchberg1974Superresolution,DeSantis1975Iterative}. Since then, many types of priors have been studied, including  Markov random fields \cite{Schultz1996Extraction}, total variation \cite{Mitzel09,Yuan2012Multiframe,Liu14}, morphological \cite{Purkait2012Super} or sparse \cite{Yang2010Image,Peleg14} models, etc. 
  Among the most effective models in the literature, many rely on the minimization of some non-differentiable functions. 
  It is for example the case of SR techniques based on sparse representations where the decomposition coefficients of the sought quantity  in a  redundant dictionary are commonly penalized by an $\ell_1$ norm, see \eg, \cite{Foucart2013Mathematical}.  
  Another example is total variation where  the $\ell_1$ norm is applied to the gradient of the sought images/motions, see \eg, \cite{Liu14}. 
  The introduction of non-differentiable functions in the SR reconstruction leads to new conundrums since standard optimization techniques for smooth problems can no longer be applied.  
  {As mentioned previously, w}e address this problem in the paper as well.  
  {Hereafter, we mainly focus on problems involving an $\ell_1$ norm, although other non-differentiable convex functions could be processed using a procedure similar to the one exposed in this paper.} 
 
{In summary, in this paper we propose a methodological framework for video SR based on a sequential model. We consider  the estimation of both a sequence of HR images and the motion fields relating them, while allowing for some non-differential terms in the cost function. Our approach is based on the combination of several modern optimization tools: fast gradient computation \cite{Bertsekas99},  the ``Alternating Direction Method of Multipliers" \cite{Boyd2011Distributed} for large-scale non-differentiable convex problems, and a recent procedure for non-convex and non-differentiable optimization proposed by Attouch \textit{et al.} in \cite{Attouch2010Proximal, Attouch2013Convergence}. The resulting algorithm is ensured to converge to a local minimum of the problem while having a linear complexity per iteration in the problem dimensions. We illustrate the good behavior of the proposed method with respect to other techniques of the state of the art in several setups. }

%

{The rest of the paper is organized as follows. We introduce the notations used throughout the paper in section \ref{sec: Notations}. In section \ref{sec:2}, we  present the sequential  model considered in our subsequent derivations. In section \ref{sec:3}, we express the video SR problem as a constrained optimization problem and provide a numerical procedure to solve it. The overall procedure is described in subsection \ref{sec:3.3} whereas two important algorithmic building blocks are presented in subsections \ref{sec:3.1} and \ref{sec:3.2}.  The numerical evaluation of the proposed method is carried out in section~\ref{sec:4} for different experimental setups.}\\
%
%
%

\section{Notations}\label{sec: Notations}

The notational conventions adopted in this paper are as follows. Italic
lowercase indicates a scalar quantity, as in $a$; boldface lowercase (resp. uppercase)
indicates a vector (resp. matrix) quantity, as in $\mathbf{a}$ (resp. $\mathbf{A}$).  
  The $n$-dimensional vector of zeroes and identity matrix will be written as $\mathbf{0}_n$ and $\mathbf{I}_n$. 
The $i$-th element
of vector $\mathbf{a}$ is denoted $\mathbf{a}(i)$;    similarly $\mathbf{A}(i,j)$ is the element of $\mathbf{A}$ located at row $i$ and column $j$. The exponent $\transpose$  denotes the transpose operation.  A  subscript notation, as in $\mathbf{a}_t$, will refer to the member of some sequence $\{\mathbf{a}_t\}_{t=0}^{T}= \{\mathbf{a}_0, \mathbf{a}_1, \cdots, \mathbf{a}_T\}$. 
 
 Calligraphic  letters, as  $\mathcal{H}$, denote functions. 
 The  subscript notation $\mathcal{H}_i$ may either denote the $i$-th element of a set $\{\mathcal{H}_i\}_{i}$ or the $i$-th component of a multidimensional function $\mathcal{H}:\Rr^m \to \Rr^n$; the distinction between these two notations is usually clear from the context. 
   The  Jacobian matrix of $\mathcal{H}:\Rr^m \to \Rr^n$ evaluated at $\tilde{\mathbf{a}}$, denoted by  $\nabla_{\mathbf{a}} \mathcal{H}(\tilde{\mathbf{a}})$,  is defined as:  
\begin{align*}
\nabla_{\mathbf{a}} \mathcal{H}(\tilde{\mathbf{a}}) =
\begin{pmatrix}
 \partial_{{\mathbf{a}}(1)} \mathcal{H}_1(\tilde{\mathbf{a}}) &\cdots& \partial_{{\mathbf{a}}(m)} \mathcal{H}_1(\tilde{\mathbf{a}})\\
\vdots & &\vdots\\
 \partial_{{\mathbf{a}}(1)} \mathcal{H}_n(\tilde{\mathbf{a}}) &\cdots& \partial_{{\mathbf{a}}(m)} \mathcal{H}_n(\tilde{\mathbf{a}})
 \end{pmatrix}\in\Rr^{n\times m},
\end{align*}
where $\partial_v$ is the partial derivative operator with respect to $v$.
 We use the notation $\nabla_{\mathbf{a}} \mathcal{H} ^*(\tilde{\mathbf{a}})$ to denote the transpose of $\nabla_{\mathbf{a}} \mathcal{H} (\tilde{\mathbf{a}})$.\\

\section{Model}\label{sec:2}

 Let $\xx_t\in\Rr^{n}$ be the image at time $t$ of an HR video sequence rearranged into a $n$-dimensional vector, with $t\in\lbrace0,\ldots,T\rbrace$. Let us suppose that we 
 capture noisy and low-resolution (LR) observations $\yy_t\in\Rr^{m}$ with $m\leq n$ of the HR sequence:
$\forall t\in\lbrace0, \ldots,T\rbrace$, 
\begin{align}
\yy_t= \mathcal{H}(\xx_t)+\boldsymbol{\eta}_t,\label{eq:y}
\end{align}
where  $\boldsymbol{\eta}_t \in \Rr^m$ stands for  some noise  and $\mathcal{H}: \Rr^{n} \rightarrow \Rr^{m}$  denotes a linear function, which is the composition of  a low-pass filtering and a sub-sampling operation. We focus on the problem of recovering the HR sequence  $\lbrace \xx_t\rbrace_{t=0}^T$ from the LR observations $\{\yy_t\}_{t=0}^T$.

Without any additional information, this problem is ill-conditioned since the number of unknowns (that is $(T+1) n$) is larger than the number of observations (that is $(T+1) m$).  One way to circumvent this problem is to take into account the relation existing between the HR images at different time instants. 
 More specifically, as part of a video, we can assume that two consecutive images obey the following  
 sequential model\footnote{ We note that \textit{backward} sequential models such as \eqref{eq:x} are common in the computer-vision literature. We therefore restrict our reasoning  to the latter formulation. However, adapting the methodologies derived in section \ref{sec:3} to a \textit{forward} sequential model is straightforward. 
}: 
 \begin{align}
\xx_t= \mathcal{P}(\xx_{t+1},\mathbf{d}_{t+1})+\boldsymbol{\epsilon}_{t+1}, \label{eq:x}
\end{align}
where  $\mathcal{P}: \Rr^{n} \times \Rr^{2n}\rightarrow \Rr^{n} $ 
is a ``warping" function characterized by a displacement $\mathbf{d}_{t+1}\in \Rr^{2n}$, and $\boldsymbol{\epsilon}_{t+1}\in \Rr^{n}$ is some noise.   The choice of $\mathcal{P}$ is usually motivated by some conservation property, as  for example the preservation  of the pixel intensity along the displacement. One particular instance of function $\mathcal{P}$, that we will consider in the sequel, is based on the well-known ``Displaced Frame Difference" model.  More specifically, this model  assumes that the $s$-th component of  $\mathcal{P}(\xx_t,\mathbf{d}_{t})$ admits the following series representation:
 \begin{align}\label{eq:Pdx}
  \mathcal{P}_s(\xx_{t},\mathbf{d}_{t})=\sum_{i\in \mathcal{V}(\chi(s)+\mathbf{d}_{t}(s))} \xx_t(i)\psi_i(\chi(s)+\mathbf{d}_{t}(s)),
 \end{align}
where $\chi:\Rr\times\Rr \to \Rr$  is a function returning the spatial position corresponding to index $s$,  $\mathcal{V}(\chi(s)+\mathbf{d}_{t}(s))$ denotes a subset of  indices corresponding to the ``neighborhood'' of point $\chi(s)+\mathbf{d}_t(s)$   and $\{\psi_i\}_{i=1}^n$ with  $\psi_i: \Rr\times\Rr \to \Rr $ is a family of  bi-dimensional polynomial interpolation functions. 
  In this case, \eqref{eq:x}-\eqref{eq:Pdx} models the fact that $\xx_{t}$ can be seen as a displaced version of $\xx_{t+1}$ plus some additive noise.  Let us note that $\mathcal{P}$,  as defined in \eqref{eq:Pdx}, is linear in $\xx_{t}$ and polynomial in $\mathbf{d}_{t}$; it is thus a bi-polynomial function. Let us also mention that $\mathcal{V}$ typically only contains a few elements that is $|\mathcal{V}|\ll n$, where $|\mathcal{V}|$ denotes the cardinality of $\mathcal{V}$; this observation will play an important role in the sequel for the analysis of the complexity of the proposed SR methodology.  

  The noise $\boldsymbol{\epsilon}_{t+1}$ in \eqref{eq:x} accounts for all the modifications of the image $\xx_t$ which cannot be inferred from $\xx_{t+1}$ and $\mathbf{d}_{t+1}$. This includes pixel occlusions, interpolation errors or variations of the scene illumination.  Notice that, in practice, the choice of $\mathcal{P}$ should be made such that the residual noise $\boldsymbol{\epsilon}_{t+1}$ is as small as possible. In particular,  if $\boldsymbol{\epsilon}_{t+1}=0$ (and $\mathbf{d}_{t+1}$ is known), $\xx_{t}$ is entirely determined from $\xx_{t+1}$ $\forall t$. Recovering the whole sequence $\lbrace \xx_t\rbrace_{t=0}^T$ is then tantamount to recovering the last image $\xx_{T}$. In such a case, the number of unknowns is therefore reduced to $n$ and the recovery of the HR sequence from the LR images may be possible. 

%
%

Another option to decrease the ill-possedness of the video  SR  problem consists in restricting the family of signals to which the ``initial condition"\footnote{We remind the reader that we consider a \textit{backward} sequential model.} $\xx_T$ belongs. 
  We will in particular\footnote{{{T}he sparsity constraints could be imposed on every $\xx_t$ without introducing any conceptual problems in the methodology exposed in section \ref{sec:3}}.} consider  the case where $\xx_T$ is assumed to be sparse in some (possibly redundant) dictionary $\mathbf{D}\in\Rr^{n\times q}$, that is
\begin{align}
\xx_T = \mathbf{D} \mathbf{c} \quad \mbox{for some $\mathbf{c} \in\Rr^{q}$ such that $\|\mathbf{c}\|_0\ll n$}, \label{eq:sparseprior}
\end{align}
where $\|.\|_0$ is the so-called ``$\ell_0$ norm'', which returns the number of nonzero coefficients of its argument. 
 Dealing with $\|.\|_0$ leads to combinatorial  optimization problems. 
 {Hereafter we will thus consider the}
$\ell_1$ norm{, which} is a well-known surrogate to the $\ell_0$ norm. 
 In particular, if the sparsity of the sought vector is large enough, there exists an equivalence between the solution of the problems involving the $\ell_0$  and $\ell_1$ norms, see \cite{Foucart2013Mathematical}.

 Finally, let us mention that the displacement $\mathbf{d}_{t}$ between two successive images is rarely known in practice. It must  therefore be inferred from the received LR images $\{\yy_t\}_{t=0}^T$. This may seem to be counterproductive since the estimation of $\mathbf{d}_{t}$ implies an increase of the number of unknowns of $2 n$ elements per time step. One way to circumvent this problem consists again in constraining $\mathbf{d}_{t}$ to belong to a restricted family of signals.  {In this paper, we consider} an implicit restriction by enforcing  some non-negative function of $\mathbf{d}_{t}$ to be small. 
 {More specifically}, we assume that the sought displacement is such that $\mathcal{R}(\mathbf{G}^* \mathbf{d}_{t})$ is ``small", 
 where $\mathbf{G}=[\mathbf{g}_1, \ldots, \mathbf{g}_h] \in \Rr^{2n\times h}$  
 is  some linear ``analysis''  operator and $\mathcal{R}: \Rr^{h}\rightarrow \Rr$ is a non-negative function.
 We note that this approach is commonly adopted in the computer-vision literature in which many options for $\mathcal{R}$ and $\mathbf{G}$ have been proposed, see \cite{Baker2011Database}. {This approach was also used in the ``multi-frame'' setting \cite{Mitzel09, Liu14} where the motions between the reference HR frame and the LR observations were penalized to have a small TV-norm.} 
In the sequel, we will focus on the following choice for $\mathbf{G}$ and $\mathcal{R}$: the elements of $\mathbf{G}^*\mathbf{d}_t$ will correspond  to the spatial gradients  of (an interpolation of) $\mathbf{d}_t$   at each point of the pixel grid; $\mathcal{R}(\mathbf{G}^*\mathbf{d}_t)$ takes the following form:
\begin{align}\label{eq:DefR}
\mathcal{R}(\mathbf{G}^*\mathbf{d}_t)\triangleq \sum_{i=1}^n \mathbf{w}(i) \Bigl|\sum_{j\in \mathcal{S}_i }(\mathbf{g}^*_j \dd_t)^2\Bigr|^{\frac{p}{2}},\quad p\ge 0,
\end{align}
  where $\mathbf{w}$ is defined as a vector of weights and $\mathcal{S}_i$'s  denote disjoint subsets of elements of $\{1,\ldots,h\}$.  Index $i$ represents a location on the pixel grid. The subset $\mathcal{S}_i$  typically gathers 4 elements corresponding to the 2 spatial gradients of the 2 components of motion $\mathbf{d}_t$ at the location indexed by $i$.
 For $p=2$, these choices are equivalent to constraining the spatial gradient of the displacement field $\mathbf{d}_t$ by a quadratic penalization \cite{Weickert04}, whereas the case  $p=1$ corresponds to the weighted total variation (TV) approach suggested  in \cite{Zach07}.
%

In summary, \eqref{eq:y}-\eqref{eq:sparseprior} together with the definition of $\mathbf{G}$ and $\mathcal{R}$ specify our prior/observation model. In the next section, we will present a low-complexity  methodology  exploiting this model to recover the HR sequence from the collected observations $\{\yy_t\}_{t=0}^T$. More specifically, we will assume that the unknowns of the problem include the HR sequence $\{\xx_t\}_{t=0}^T$, the sequential noise $\{\epsi_t\}_{t=1}^T$, the displacements $\{\mathbf{d}_t\}_{t=1}^T$ and the decomposition vector $\mathbf{c}$. All the other parameters of the problem will be supposed to be known, although they could  easily be included as additional unknowns without introducing any conceptual problem in the proposed methodology. \\

\section{The Estimation Procedure}\label{sec:3}
In this section, we expose our methodology to estimate the HR sequence by exploiting the model described in section \ref{sec:2}. Our approach is based on the resolution of a constrained optimization problem.  We introduce the following shorthand notations: 
  $\xx\triangleq\{\xx_t\}_{t=0}^T $, $\epsi \triangleq\lbrace \epsi_t \rbrace_{t=1}^{T}$ and $\dd \triangleq\lbrace \dd_t \rbrace_{t=1}^{T}$. 
 Our SR reconstruction procedure relies on  the  following constrained optimization problem:
\begin{equation}\label{eq:prob2}
 \argmin_{(\xx,\epsi,\dd, \mathbf{c})}  \mathcal{J}(\xx,\epsi,\dd,\mathbf{c}) 
\quad \textrm{s.t.}\ \left\{\begin{aligned}
&\xx_t=\mathcal{P}(\xx_{t+1},\dd_{t+1})+\boldsymbol{\epsilon}_{t+1},&\hspace{0.2cm} 0\leq t \leq  T-1, \\
&\xx_T=\mathbf{D} \cv,&\hspace{0.2cm}
  \end{aligned}\right.
\end{equation}
where 
\begin{align*}
  \mathcal{J}(\xx,\epsi,\dd,\mathbf{c})\triangleq  
  & \sum_{t=0}^{T}\| \mathcal{H}( \xx_t) -\yy_t \|^2_2
  + \alpha_1 \sum_{t=1}^{T} \|  \epsi_t  \|_p^p  
  + \alpha_2 \sum_{t=1}^{T}  \mathcal{R}( \mathbf{G}^*\dd_t  )
  + \alpha_3\| \cv\|^p_p,
\end{align*}
for some $\alpha_j\geq 0$, $j\in\{1,2,3\}$, and $p\geq 0$. 
Let us make a few comments about \eqref{eq:prob2}. The first constraint 
 ensures that the images of the HR sequence verify the sequential model \eqref{eq:x}; the second enforces that prior model \eqref{eq:sparseprior} is satisfied. Each term in the cost function $\mathcal{J}(\xx,\epsi,\dd,\mathbf{c})$ has a clear physical meaning: the first term penalizes the discrepancies between the predicted and the received observations; the second penalizes the noise on the sequential model; the third enforces the displacement to have some regularity and the last one constrains $\mathbf{c}$ to have some desirable properties  (depending on the choice of $p$). For example, setting $p\in[0,1]$ typically promotes the sparsity of $\mathbf{c}$, see \cite{Foucart2013Mathematical}. {Because sparsity has revealed to be a good prior in a number of works, in the following our main objective is to find a solution to \eqref{eq:prob2} with $p=1$}. 
%


Problem  \eqref{eq:prob2} involves a huge number of unknowns (namely $(4T+1)n+q$ variables if $\xx,\epsi,\dd,\mathbf{c}$ have to be estimated).  Hence, solving \eqref{eq:prob2} may be critical even for reasonable problem sizes: for instance, considering images of $n=2^8 \times 2^8$ pixels, a non-redundant dictionary, \ie, $q=n$ and a sequence length $T=2^4$, we have that the number of variables involved in the optimization problem grows up to roughly $2^{22}$. Clearly, such a high-dimensional problem can only be addressed by specifically-dedicated procedures. 
 {In subsection \ref{sec:3.3}, we propose an overall methodology to solve \eqref{eq:prob2} efficiently with $p=1$. 
 Our approach is based on the combination of several modern optimization tools, described in subsections \ref{sec:3.1} and \ref{sec:3.2}. 
 More specifically, the building blocks presented in subsections \ref{sec:3.1} and \ref{sec:3.2} tackle simplified versions of problem \eqref{eq:prob2}, which appear as intermediate steps in the overall procedure described in section \ref{sec:3.3}. We briefly comment on these intermediate problems  in the next paragraphs.}
 

{In section \ref{sec:3.1}, we consider the case where $p=2$ in \eqref{eq:prob2}, that is all the functions are differentiable.} 
 In such a case, we show that the gradient of the cost function associated to an (equivalent) unconstrained version of  \eqref{eq:prob2} can be evaluated efficiently by resorting to  optimal control techniques  \cite{Bertsekas99}.   
 More specifically, we emphasize that the complexity  associated to the evaluation of the gradient of the cost function remains \textit{linear} in the problem dimensions, for many setups of practical interest.  

In section \ref{sec:3.2}, we  {focus on} the case where $\dd$ is known 
 but  $p=1$. The corresponding optimization problem is then convex but  not differentiable. Building on our derivations in section \ref{sec:3.1}, we emphasize that this type of problem can be nicely addressed by resorting to the so-called ``Alternating Direction of Multipliers Method"  (ADMM) \cite{Boyd2011Distributed}, a modern optimization technique proposed to handle large-scale non-differentiable optimization problems. 


Finally, in section \ref{sec:3.3}, we consider the general problem \eqref{eq:prob2}, where $\xx,\epsi,\dd,\mathbf{c}$ have to be estimated and $p=1$. In this case, \eqref{eq:prob2} is non-convex (because the term $\mathcal{P}(\xx_{t+1},{\dd_{t+1}})$ appearing in the constraints is bi-polynomial) and  non-differentiable. In order to address this problem, we  resort to an optimization procedure introduced by Attouch \textit{et al.} \cite{Attouch2010Proximal,Attouch2013Convergence}, and particularized in \cite{Puy2014Robust} to multi-frame SR. The procedure is iterative and exploits the building blocks derived in sections \ref{sec:3.1} and \ref{sec:3.2} {to solve intermediate problems}. The complexity per iteration  is  linear in the problem dimensions. Moreover, from the arguments exposed in \cite{Puy2014Robust},  it can be shown that the proposed procedure is convergent to a critical point of the problem. \\


%
%
\subsection{{The First Building Block}}\label{sec:3.1}

In this section, we assume {that $p=2$ (so} that all the functions appearing in \eqref{eq:prob2} are differentiable) and show that an efficient resolution of  \eqref{eq:prob2} via gradient descent algorithms exists. Our approach is based on fast gradient evaluation techniques as exposed in \cite{Bertsekas99}. 

In order to present our methodology, we first reformulate \eqref{eq:prob2} as an (equivalent) unconstrained problem. Notice that, because of the constraints in problem  \eqref{eq:prob2}, any $\xx_t$ can be expressed as a deterministic function of $\mathbf{c}$ and $\{ \epsi_{t'},\dd_{t'}\}_{t'=t+1}^{T}$. In other words, there exists a function $\mathcal{Q}(\epsi,\dd,\mathbf{c}):\Rr^{Tn}\times\Rr^{2Tn}\times\Rr^q\rightarrow \Rr^{(T+1)n}$ such that, given $\epsi,\dd$ and $\mathbf{c}$,  $\xx=\mathcal{Q}(\epsi,\dd,\mathbf{c})$ is the unique vector satisfying the constraints in \eqref{eq:prob2}. As a consequence, \eqref{eq:prob2} can also be equivalently expressed as
\begin{align}
 \argmin_{(\epsi,\dd, \mathbf{c})}  \mathcal{J}(\epsi,\dd,\mathbf{c}),\label{eq:unprob}
\end{align}
where 
\begin{align}
  \mathcal{J}(\epsi,\dd,\mathbf{c}) \nonumber
  &\triangleq \mathcal{J}(\xx=\mathcal{Q}(\epsi,\dd,\mathbf{c}),\epsi,\dd,\mathbf{c}),\\
  & =\sum_{t=0}^{T}\| \mathcal{H}( \mathcal{Q}_t(\epsi,\dd,\mathbf{c})) -\yy_t \|^2_2
   + \alpha_1 \sum_{t=1}^{T} \|  \epsi_t  \|_p^p  \nonumber\\
  &+ \alpha_2 \sum_{t=1}^{T}  \mathcal{R}( \mathbf{G}^*\dd_t  )
  + \alpha_3\| \cv\|^p_p,\label{eq:costfunctionun}
\end{align}
and $\mathcal{Q}_t(\epsi,\dd,\mathbf{c})$ is the restriction of $\mathcal{Q}(\epsi,\dd,\mathbf{c})$ to $\xx_t$. 

Since {$p=2$}, \eqref{eq:unprob} is a smooth \textit{unconstrained} minimization problem and can thus be solved by any procedure belonging to the family of gradient descent algorithms.  At this point, let us make two remarks: 
 \textit{i)} $\mathcal{J}(\epsi,\dd,\mathbf{c})$ has usually an intricate structure and its gradient does therefore not have any simple analytical expression;  \textit{ii)} the computation of the gradient of $\mathcal{J}(\epsi,\dd,\mathbf{c})$ via finite differences is out of reach for the considered problem because  it would require to evaluate the cost function twice as many times as the (huge!) number of variables. 
 
 As a consequence, the main bottleneck for solving  \eqref{eq:unprob}  lies in the tractable  evaluation of the gradient of $\mathcal{J}(\epsi,\dd,\mathbf{c})$.
 We emphasize in appendix \ref{sec:app1} that the particular structure of $\mathcal{J}(\epsi,\dd,\mathbf{c})$  enables the use of a specific methodology   with a complexity scaling linearly with the problem dimensions. More specifically, let
\begin{equation}\label{eq:defG0}
\left\{\begin{aligned}
\mathcal{G}_0(\xx_0) & \triangleq \| \mathcal{H}( \xx_0)-\yy_0 \|^2_2,\\
\mathcal{G}_t(\xx_t,\epsi_t,\dd_t) & \triangleq \| \mathcal{H}( \xx_t) -\yy_t \|^2_2+ \alpha_1 \|  \epsi_t  \|^p_p+ \alpha_2 \mathcal{R}( \mathbf{G}^*\dd_t  ) ,\hspace{0.3cm}\text{for}\;\; 1\leq t \leq T-1,\\
\mathcal{G}_T(\xx_T,\epsi_T,\dd_T,\cv) & \triangleq \| \mathcal{H} (\xx_T) -\yy_T \|^2_2 + \alpha_1 \|  \epsi_T  \|^p_p+ \alpha_2 \mathcal{R}( \mathbf{G}^*\dd_T  ) + \alpha_3\| \cv\|^p_p.
  \end{aligned}\right.
\end{equation}
 Using the notation $\mathcal{G}_T(\epsi_T,\dd_T,\cv) \triangleq \mathcal{G}_T(\xx_T=\mathbf{D}\cv,\epsi_T,\dd_T,\cv) $,
the elements of the gradient of $\mathcal{J}(\epsi,\dd,\mathbf{c})$ at $(\epsi',\dd',\mathbf{c}')$ can then be evaluated as follows: 
\begin{equation}\label{eq:grad1}
\left\{
\begin{aligned}
&\nabla_{\dd_{t}} \mathcal{J}(\epsi',\dd',\mathbf{c}')=\nabla_{\dd_{t}} \mathcal{P}^*(\xx_{t}',\dd_t') \zv_{t-1}+ \nabla_{\dd_t}  \mathcal{G}_{t}(\xx_{t}',\epsi_t',\dd_t'),\\
&\nabla_{\boldsymbol{\epsilon}_{t}} \mathcal{J}(\epsi',\dd',\mathbf{c}')= \zv_{t-1}+ \nabla_{\boldsymbol{\epsilon}_{t}}  \mathcal{G}_{t}(\xx_{t}',\epsi_t',\dd_t'),\\
&\nabla_{\cv} \mathcal{J}(\epsi',\dd',\mathbf{c}')=\mathbf{D}^* \zv_T+ \nabla_{\cv}  \mathcal{G}_{T}(\epsi_T',\dd_T',\cv'),
\end{aligned}\right.
\end{equation}
where the variables $\xx_t'$, $\epsi_t',\dd_t'$ and $\mathbf{c}'$ must satisfy the constraints of problem \eqref{eq:prob2}, that is 
\begin{align} \label{eq:direct}
\left\{\begin{aligned}
&\xx_T'=\mathbf{D}\cv',&\hspace{0.2cm}\\
&\xx_t'= \mathcal{P}(\xx'_{t+1},\mathbf{d}'_{t+1})+\boldsymbol{\epsilon}'_{t+1}, &\hspace{0.2cm} t=T-1,...,0, 
  \end{aligned}\right.
\end{align}
and the sequence of ``adjoint" variables $\{\zv_t\}_{t=0}^T$ obeys the following recursion:
 \begin{equation}\label{eq:foward1}
\left\{\begin{aligned}
&\zv_0 =\nabla \mathcal{G}_0(\xx_0'), \\
&\zv_{t}=\nabla_{\xx_{t}} \mathcal{P}^*(\xx_{t}',\dd_t') \zv_{t-1} + \nabla_{\xx_{t}}  \mathcal{G}_{t}(\xx_{t}',\epsi_t',\dd_t'), \hspace{0.3cm} t=1,...,T-1,\\
&\zv_T=\nabla_{\xx_{T}} \mathcal{P}^*(\xx_{T}',\dd_T') \zv_{T-1} + \nabla_{\xx_{T}}  \mathcal{G}_{T}(\xx_T',\epsi_T',\dd_T',\cv').
\end{aligned}\right.
\end{equation}
Expressions in \eqref{eq:grad1} together with recursions  \eqref{eq:direct} and \eqref{eq:foward1} provide an efficient way to evaluate  the gradient of $\mathcal{J}(\epsi,\dd,\mathbf{c})$. The overall methodology can be understood as a 3-step procedure: \textit{i)} given some values of $\epsi',\dd'$ and $\mathbf{c}'$, evaluate $\{\xx_t'\}_{t=0}^T$ with recursion \eqref{eq:direct}; \textit{ii)} use the value of $\{\xx_t'\}_{t=0}^T$ to evaluate the adjoint variables $\{\zv_t\}_{t=0}^T$ from \eqref{eq:foward1}; \textit{iii)} compute the gradient of $\mathcal{J}(\epsi,\dd,\mathbf{c})$ by using \eqref{eq:grad1}.
 Note that the gradients appearing in the right-hand side of \eqref{eq:grad1} and \eqref{eq:foward1} typically have simple analytical expressions and are thus straightforward to evaluate. 

 It is  easy to see that the  complexity induced by this methodology scales (at worst) as $\mathcal{O}(n^2 T + nq)$ since it only involves matrix-vector multiplications, with matrices of dimension $n \times n $ or $n \times q$. In practice, this complexity can usually be reduced to $\mathcal{O}(n T+q)$, or simply  to $\mathcal{O}(n T)$ in the case of a non-redundant dictionary. This  linearity in the problem dimensions occurs if the matrices involved in \eqref{eq:sparseprior}, \eqref{eq:grad1} and \eqref{eq:foward1}  are typically very sparse and/or  rely on fast transforms of  linear complexity\footnote{This is the case for any non-redundant wavelet basis, which will induce an overall complexity of $\mathcal{O}(nT)$. Fast transforms for sparse redundant dictionaries such as curvelets frames  also exist but imply a slight complexity overload since the matrix-vector multiplication scales in this case as $\mathcal{O}(n \log n)$, yielding an overall complexity of  $\mathcal{O}(n(T+ \log n))$.}.  In the (typical) example \eqref{eq:Pdx} considered in this paper, we clearly obtain this linear complexity since $|\mathcal{V}(\chi(s)+\dd_t(s))|\ll n$. 
  In the rest of the paper, we focus on  model \eqref{eq:Pdx} and  choose a dictionary so that the complexity  related to \eqref{eq:grad1}-\eqref{eq:foward1} is linear in the problem dimensions. 
%

 {Before concluding this section, let us open a parenthesis to highlight some connections with some previous works which considered the ``Kalman smoother'' update rules as the starting point of their video SR method, see \cite{Elad1999Superresolution}.} 
 {First notice that, assuming $\dd$  is known, \eqref{eq:unprob} with $p=2$ corresponds to the ``Maximum a Posteriori''  
  (MAP) estimation problem associated to the following probabilistic (backward) state-evolution model:}
\begin{align}
\left\lbrace
\begin{array}{ll}
\xx_{T}& \sim \mathcal{N}(\mathbf{0}_n, \alpha_3^{-1} \mathbf{D}\mathbf{D}^*)\\
 \xx_{t} &\sim \mathcal{N}(\mathcal{P} (\xx_{t+1},\mathbf{d}_{t+1}),\alpha_1^{-1} \mathbf{I}_n )\\
 \yy_{t} &\sim \mathcal{N}(\mathcal{H}(\xx_t),\mathbf{I}_m),
\end{array}
\right.
\end{align}
where $\mathbf{v} \sim  \mathcal{N}(\mathbf{m},\Gamma)$ indicates  that $\mathbf{v}$ is distributed according to a multivariate normal distribution with mean $\mathbf{m}$ and covariance $\Gamma$.

 For such a model, it is well-known that the Kalman smoother can compute exactly the solution of \eqref{eq:unprob} in a finite number of steps, namely one forward and one backward recursions, see \eg, \cite[Chapter 20]{Mendel_book}. The Kalman smoother involves  the update of a length-$n$ mean vector and an $n\times n$  covariance matrix at each step of the two recursions; moreover, the evaluation of these quantities requires the inversion of a $n\times n$ matrix. Hence, the Kalman smoother exhibits a computational complexity scaling as\footnote{We note that the complexity can be reduced to $\mathcal{O}(T (m^3+mn))$ by using some computational tricks as the well-known Woodbury matrix identity, see \eg,  \cite[Lemma 4.1]{Mendel_book}. However, the latter still remain too costly for typical problem sizes.  } $\mathcal{O}(n^3 T)$. {Since this complexity is prohibitive for most practical setups, several approximations of the Kalman updates for video SR have been proposed in \cite{Elad1999Superresolution}. }
{On the other hand, the procedure described in this section provides an alternative, approximation-free, solution to the MAP problem. 
 Indeed, since  \eqref{eq:unprob} with $p=2$ is a differentiable problem, it can be solved with a simple gradient descent method. 
 More specifically, we can apply the methodology described in this section to efficiently compute the gradient of $\mathcal{J}(\epsi,\dd,\mathbf{c})$ with respect to $\epsi$ and $\mathbf{c}$ (using the two last rows of \eqref{eq:grad1}). 
 The complexity of this method then only scales as $\mathcal{O}(n T)$ per iteration. 
 Moreover, because $\mathcal{J}(\epsi,\dd,\mathbf{c})$ is strictly convex in $(\epsi,\mathbf{c})$, this type of algorithm is ensured to converge to the global minimum of the problem. 
 Hence, if the descent algorithm has converged (close) to the minimum after a reasonable number of iterations, the proposed methodology drastically reduces the complexity necessary to obtain the MAP solution as compared to a Kalman smoother.} \\



\subsection{{The Second Building Block}}\label{sec:3.2}


{In this section, we address problem \eqref{eq:prob2} with $p=1$ but assume that $\dd$ is known (and thus therefore no longer appear as an optimization variable in \eqref{eq:prob2}). }
%
 Particularizing \eqref{eq:prob2} to these working hypotheses, we obtain the following convex but non-differentiable problem:

\begin{align}\label{eq:prob_cnvexnondiff}
\argmin_{(\xx,\epsi, \mathbf{c})}  \mathcal{J}(\xx,\epsi,\mathbf{c}) 
\quad \textrm{s.t.}\ \left\{
\begin{array}{ll}
\xx_t=\mathcal{P}(\xx_{t+1},\dd_{t+1})+\epsi_{t+1},&\hspace{0.2cm} 0\leq t \leq  T-1 \\
\xx_T=\mathbf{D} \cv,&\hspace{0.2cm}
 \end{array}
  \right.
\end{align}
where
\begin{align*}
  \mathcal{J}(\xx,\epsi,\mathbf{c})\triangleq  
  & \sum_{t=0}^{T}\| \mathcal{H} (\xx_t) -\yy_t \|^2_2
  + \alpha_1 \sum_{t=1}^{T} \|  \epsi_t  \|_1  
  + \alpha_3\| \cv\|_1.
\end{align*}
{This problem is convex but non-differentiable.}  
 As previously, the main bottleneck for its resolution lies in its high dimensionality. This, in turn, imposes to resort to low-complexity optimization procedures. We show hereafter, that a complexity scaling linearly with the problem dimensions is possible by  using the ``Alternating Direction Method of Multipliers'' (ADMM). ADMM has recently emerged in the optimization community to address large-scale optimization problems. Among the particular assets of this type of method, let us mention: \textit{i)} its robustness (the convergence to a global minimum is ensured under very mild conditions);  \textit{ii)} its rapid convergence to  an acceptable accuracy (typically a few tens of iterations is sufficient). We refer the reader to appendix \ref{app:ADMM} for a short description of the ADMM framework. 

In order to derive the ADMM recursions, we first need to reformulate \eqref{eq:prob_cnvexnondiff} in the standard form \eqref{eq:ADMMstdproblem} in appendix \ref{app:ADMM}. Letting
\begin{align}\label{eq:setC}
\Cc  \triangleq\left\{ (\xx,\epsi, \mathbf{c}) \left| 
\begin{aligned}
\xx_t&=\mathcal{P}(\xx_{t+1},\dd_{t+1})+ \epsi_{t+1},&\hspace{0.2cm} 0\leq t \leq  T-1 \\
\;\;\xx_T\negmedspace&=\mathbf{D} \cv&
\end{aligned}
\right.
\right\},
\end{align}
\eqref{eq:prob_cnvexnondiff} can be reexpressed as:
\begin{align}\nonumber
\argmin_{\substack{(\xx,\epsi, \mathbf{c})\in\Cc\\ (\tilde{\epsi}, \tilde{\mathbf{c}})\in\Rb^{nT+q}}} \quad&\sum_{t=0}^{T}\| \mathcal{H} (\xx_t )-\yy_t \|^2_2
  + \alpha_1 \sum_{t=1}^{T} \|  \tilde{\epsi}_t  \|_1  
  + \alpha_3\| \tilde{\cv}\|_1,\\
 & \textrm{s.t.}\ \left\{ \nonumber
\begin{array}{ll}
\epsi=\tilde{\epsi},\\
\mathbf{c}=\tilde{\mathbf{c}}.\\
 \end{array}
  \right.
\end{align}
Here, we have added two new variables to the problem, $\tilde{\epsi}$ and $\tilde{\mathbf{c}}$, which are counterbalanced by the inclusion of two new constraints. Using the formalism exposed in appendix \ref{app:ADMM} with $\zz_1=(\xx,\epsi, \mathbf{c})$, $\zz_2=(\tilde{\epsi}, \tilde{\mathbf{c}})$, $\Zc_1=\Cc$ and  $\Zc_2=\Rb^{nT+q}$, we obtain the following ADMM recursions:
\begin{align}
&(\xx^{(k+1)}, \epsi^{(k+1)}, \mathbf{c}^{(k+1)}) = \argmin_{\xx,\epsi,\mathbf{c}}\mathcal{L}^{(k)} (\xx,\epsi,\mathbf{c})\label{eq:step1_ADMM}\\
 &\quad\quad\textrm{s.t.}\ \left\{ \nonumber
\begin{aligned}
\xx_t&=\mathcal{P}(\xx_{t+1},\dd_{t+1})+\epsi_{t+1},&\hspace{0.2cm} 0\leq t \leq  T-1 \\
\xx_T&=\mathbf{D} \cv,&\hspace{0.2cm}
 \end{aligned}
  \right.\\
 & 
 \left\{
\begin{array}{ll}
\tilde{\epsi}^{(k+1)}_t =& \argmin_{\tilde{\epsi}_t}\  \|  \tilde{\epsi}_t  \|_1 + \frac{\rho_1}{2 \alpha_1} \| \epsi^{(k+1)}_t- \tilde{\epsi}_t   + \mathbf{u}_{\epsi_t}^{(k)} \|^2_2, \label{eq:step2_ADMM} \\
\tilde{\mathbf{c}}^{(k+1)}=& \argmin_{\tilde{\mathbf{c}}}\quad \| \tilde{\cv}\|_1+ \frac{\rho_3}{2 \alpha_3} \|  \mathbf{c}^{(k+1)} - \tilde{\mathbf{c}} + \mathbf{u}_{\mathbf{c}}^{(k)} \|^2_2,\\
\end{array}
 \right.\\
 &
 \left\{
 \begin{array}{ll}
\mathbf{u}_{\epsi_t}^{(k+1)} =& \mathbf{u}_{\epsi_t}^{(k)} + \epsi^{(k+1)}_t - \tilde{\epsi}^{(k+1)}_t, \label{eq:step3_ADMM}\\
\mathbf{u}_{\mathbf{c}}^{(k+1)}=& \mathbf{u}_{\mathbf{c}}^{(k)} + \mathbf{c}^{(k+1)}- \tilde{\mathbf{c}}^{(k+1)}, 
\end{array}
 \right.
\end{align}
where $\rho_1,\rho_3>0$ and we have introduced  the function:
\begin{align}\label{eq:Lk}
\mathcal{L}^{(k)} (\xx,\epsi,\mathbf{c})& = \sum_{t=0}^{T}\| \mathcal{H} (\xx_t) -\yy_t \|^2_2+ \frac{\rho_1}{2}\| \epsi - \tilde{\epsi}^{(k)} + \mathbf{u}_{\epsi}^{(k)} \|^2_2 
 + \frac{\rho_3}{2}\| \mathbf{c} - \tilde{\mathbf{c}}^{(k)} + \mathbf{u}_{\mathbf{c}}^{(k)} \|^2_2 .
\end{align}
Equations \eqref{eq:step1_ADMM}, \eqref{eq:step2_ADMM} and \eqref{eq:step3_ADMM} correspond respectively to expressions \eqref{eq:ADMMstdformA}, \eqref{eq:ADMMstdformB} and \eqref{eq:ADMMstdformC} in appendix \ref{app:ADMM}. 

Let us make the following remarks about the different steps of the ADMM procedure. 
First, problem \eqref{eq:step1_ADMM} has the same structural form as the problem addressed in section \ref{sec:3.1}; in particular, all the terms of the cost function appearing in  \eqref{eq:step1_ADMM} are differentiable while the set of constraints imposed on $\xx,\epsi,\dd$ and $\mathbf{c}$ is strictly the same. We can thus apply the methodology described in section \ref{sec:3.1} to solve this problem via a gradient descent algorithm, with a complexity per iteration scaling as $\mathcal{O}(nT)$. Interestingly, let us mention that, under very mild conditions, the convergence of ADMM is still guaranteed if the minimizations in \eqref{eq:step1_ADMM}-\eqref{eq:step2_ADMM} are not  performed exactly, see \eg,  \cite[Theorem 8]{Eckstein1992DouglasRachford}. This suggests that the number of gradient steps carried out to search for the minimum of \eqref{eq:step1_ADMM} can be rather limited without affecting the convergence of the overall ADMM process. 

Second, the optimization problems specified in \eqref{eq:step2_ADMM} have a very simple analytical solution. In fact the right-hand sides of \eqref{eq:step2_ADMM} correspond to the definition of the proximal operator of the $\ell_1$ norm. The latter has been extensively studied in the literature (see \eg, \cite[section 6.5.2]{OPT-003}) and possesses a simple analytical solution based on soft-thresholding operators. In particular, we have 
\begin{align}\label{eq:solstep2}
\left\{
\begin{array}{ll}
\tilde{\epsi}^{(k+1)}_t(i)&=\mathrm{soft}_{\frac{\alpha_1}{\rho_1}} \left(\epsi^{(k+1)}_t(i) + \mathbf{u}_{\epsi_t}^{(k)}(i)\right)\\
\tilde{\mathbf{c}}^{(k+1)}(i)&=\mathrm{soft}_{\frac{\alpha_3}{\rho_3}} \left( \mathbf{c}^{(k+1)}(i) + \mathbf{u}_{\mathbf{c}}^{(k)}(i) \right)
\end{array}
\right. \qquad\forall i,
\end{align}
where
\begin{align} \label{eq:softthresh}
\mathrm{soft}_{\lambda}\left( a\right)&=
\left\{
\begin{array}{ll}
a- \lambda & \mbox{if $a \geq \lambda$,}\\
a+ \lambda & \mbox{if $a\leq - \lambda$,}\\
0 & \mbox{otherwise.}
\end{array}
\right. 
\end{align}
We note that the solution of \eqref{eq:step2_ADMM} is typically sparse since the soft-thresholding operator \eqref{eq:softthresh} enforces the small coefficients to be equal to zero. 
 Moreover, we see from  \eqref{eq:solstep2} that the complexity of this ADMM step clearly  scales as $\mathcal{O}(nT+q)$. 


As a conclusion, since the last step \eqref{eq:step3_ADMM} of the procedure only involves vector additions, the particularization of ADMM to our problem leads to an algorithm exhibiting a complexity per iteration scaling linearly in the problem dimensions. \\

\subsection{The Overall Procedure}\label{sec:3.3}
Let us now concentrate our attention on our target problem, that is \eqref{eq:prob2} with $p=1$, where all the variables $\mathbf{x}$, $\mathbf{c}$, $\epsi$, $\dd$ have to be estimated. The cost function then contains both non-differentiable and non-convex terms. In such a case, ensuring the convergence to a global minimum is usually out of reach for any deterministic optimization procedure. In this section, we consider an optimization method proposed in \cite{Attouch2010Proximal,Attouch2013Convergence} and particularized  to multi-frame super-resolution problems in \cite{Puy2014Robust}. This procedure addresses optimization problems involving a cost function satisfying the so-called ``Kurdyca-Lojasiewicz'' property and is guaranteed to convergence to a critical point of the latter under mild conditions. We refer the reader to \cite{Attouch2010Proximal,Attouch2013Convergence} for more details about ``Kurdyca-Lojasiewicz" functions. 
 Here, we just mention that functions made up of the composition of piecewise polynomial functions obey the ``Kurdyca-Lojasiewicz" property. Scrutinizing the structure of \eqref{eq:prob2} and taking \eqref{eq:Pdx} into account, it is easy to see that our cost function is piecewise polynomial; the optimization framework developed in \cite{Attouch2010Proximal,Attouch2013Convergence} therefore applies.

 

Our methodology obeys a 2-step recursion which follows the same lines as the procedure presented in \cite{Puy2014Robust}. {The building blocks described in subsections \ref{sec:3.1} and \ref{sec:3.2} are used to provide an efficient implementation of the intermediate problems appearing in these two steps.} To express the procedure recursions, we focus on the unconstrained formulation \eqref{eq:unprob} {(with $p=1$)} of our general optimization problem \eqref{eq:prob2}. 
The first step of the procedure solves the following problem: 
\begin{align}
(\epsi^{(k+1)}, \mathbf{c}^{(k+1)}) 
&= \argmin_{(\epsi, \mathbf{c})} \mathcal{J}(\epsi, \mathbf{d}^{(k)}, \mathbf{c})+\gamma\, \mathcal{C}(\epsi- \boldsymbol{\epsilon}^{(k)},\cv-  \cv ^{(k)}),\label{eq:Puy1} 
\end{align}
 where $\gamma>0$,  $\mathcal{J}$ is the cost function in \eqref{eq:costfunctionun}   {with $p=1$} and $\mathcal{C}:\mathbb{R}^{nT}\times \mathbb{R}^q \rightarrow \mathbb{R}_+$  is a non-negative proper lower-semicontinuous convex function such that $\mathcal{C}(\mathbf{0}_{nT},\mathbf{0}_q)=0$. It thus consists in minimizing the (penalized) cost function  $\mathcal{J}(\epsi, \mathbf{d},\mathbf{c})$ over the subset of variables $(\epsi, \mathbf{c})$;   the  penalizing term $\mathcal{C}$  plays the role of a ``cost-to-move" function which prevents the new iterate $(\epsi^{(k+1)}, \mathbf{c}^{(k+1)})$ from differing too much from the previous one. In the sequel, we will focus on the following penalizing term\footnote{In theory, the $\ell_1$-norm should be substituted by a smooth approximation to prove convergence towards a critical point of the cost function, as done in  \cite{Puy2014Robust}. In practice, we note that this substitution does not impact convergence.}: 
\begin{align}
\mathcal{C}(\epsi- \boldsymbol{\epsilon}^{(k)},\cv-  \cv ^{(k)})&= \sum_{t=1}^T \| \mathbf{H}^* (\boldsymbol{\epsilon}_t- \boldsymbol{\epsilon}_t^{(k)})\|_1+  \| \cv-  \cv ^{(k)}\|_1, \label{eq:cost2move}
\end{align}
where $\mathbf{H} \in \Rr^{n \times n}$ is a wavelet basis. The operational meaning of this ``cost-to-move" function is as follows: the $\ell_1$ norm enforces its argument to be sparse; hence, the second term in \eqref{eq:cost2move} ensures that the number of nonzero coefficients in $\mathbf{c}^{(k+1)}$ does not differ too much from the one in $\mathbf{c}^{(k)}$, while the first term plays the same role for the wavelet coefficients of $\boldsymbol{\epsilon}_t^{(k+1)}- \boldsymbol{\epsilon}_t^{(k)}$. Using this type of ``cost-to-move'' is not mandatory for the convergence of the proposed procedure. However, it has been shown empirically in \cite{Puy2014Robust} that it is well-suited to avoid some undesirable local minima of the cost function\footnote{An intuitive explanation is that the cost-to-move \eqref{eq:cost2move} induces a ``coarse-to-grain'' refinement  of the unknowns which is usually beneficial in computer-vision problems, see details in \cite{Puy2014Robust}.}.

In the second step of the recursion, we update the velocity field $\mathbf{d}$ as:
\begin{align}
\mathbf{d}^{(k+1)} 
&=  \argmin_{\mathbf{d}} \mathcal{B}(\mathbf{d},\mathbf{d}^{(k)})
  + \alpha_2 \sum_{t=1}^{T}  \mathcal{R}( \mathbf{G}^*\dd_t  ), \label{eq:Puy2}
\end{align}
where $\mathcal{B}(\mathbf{d},\mathbf{d}^{(k)})$ is a quadratic approximation\footnote{Note that $\mathcal{B}(\mathbf{d})$ is similar to the first term of the cost function in \eqref{eq:costfunctionun}.} 
 of $ \mathcal{B}(\mathbf{d})\triangleq \sum_{t=0}^{T}\| \mathcal{H}( \mathcal{Q}_t(\epsi^{(k+1)},\dd,\mathbf{c}^{(k+1)})) -\yy_t \|^2_2$, that is
 \begin{align}
\mathcal{B}(\mathbf{d},\mathbf{d}^{(k)}) &\triangleq \mathcal{B}(\mathbf{d}^{(k)})+   \nabla_{\dd} \mathcal{B}^*(\mathbf{d}^{(k)})(\mathbf{d}-\mathbf{d}^{(k)})  + \frac{\alpha^{(k)}}{2}\|  \mathbf{d}-\mathbf{d}^{(k)}\|^2_2, \quad \alpha^{(k)}>0.\label{eq:defquadapprox}
\end{align}
The choice of $\alpha^{(k)}$ is of course not arbitrary and should be made so that the convergence of the procedure is ensured. We elaborate on this point further in this section. For now, let us first discuss the practical implementation and complexity of recursions \eqref{eq:Puy1}-\eqref{eq:Puy2}.  
 It should be noticed that the building blocks presented in sections \ref{sec:3.1} and \ref{sec:3.2} can be exploited to solve efficiently these steps. 
Indeed, problem \eqref{eq:Puy1} has the same structural form as the one considered in \eqref{eq:prob_cnvexnondiff}: the cost function consists in a quadratic term plus a set of convex but non-differentiable terms. 
We can thus use the ADMM procedure described in section \ref{sec:3.2} to address it. In the same way, we see from definition \eqref{eq:defquadapprox} that the cost function \eqref{eq:Puy2} is made up of a quadratic term plus some non-differentiable function $\alpha_2 \sum_{t=1}^{T}  \mathcal{R}( \mathbf{G}^*\dd_t )$. Hence, the ADMM procedure described in  section \ref{sec:3.2} can also be applied here to solve \eqref{eq:Puy2}. 
 In comparison to our exposition in  section \ref{sec:3.2}, only the proximal operators of the non-differentiable terms will change when ADMM is applied to \eqref{eq:Puy1} and \eqref{eq:Puy2}.  In particular, the computation of any gradient of the differentiable part of the cost functions can be efficiently evaluated via the procedure described in section \ref{sec:3.1}. We particularize the expression of the proximal operators appearing in the ADMM implementation of \eqref{eq:Puy1}-\eqref{eq:Puy2}  in appendix \ref{app:attouch}. As previously, it turns out that the implementation of the latter only requires a linear complexity.  The complexity of each iteration of \eqref{eq:Puy1}, \eqref{eq:Puy2}  is thus once again linear. 
 

To conclude this section, let us discuss the convergence of the proposed procedure. In \cite[Theorem 1]{Puy2014Robust}, the authors proved that if  $\mathcal{J}(\epsi,  \mathbf{d},\mathbf{c})$  satisfies the ``Kurdyca-Lojasiewicz''  property and the $\alpha^{(k)}$'s are properly selected, the sequence defined in  \eqref{eq:Puy1}-\eqref{eq:Puy2} is either unbounded or converges to a critical point of $\mathcal{J}(\epsi,  \mathbf{d},\mathbf{c})$. 
 A procedure to properly select factors $\alpha^{(k)}$ is exposed in \cite[section 2.3]{Puy2014Robust} and is easy to implement in practice. Particularized to the setup considered in this paper, this procedure reads {as follows}: select  $\alpha^{(k)}=2^i \xi$  with $\xi>0$  and with $i$  the smallest  positive integer  such that 
 \begin{align}\label{eq:stepfactor}
\mathcal{B}(\mathbf{d}^{(k+1)})-\mathcal{B}(\mathbf{d}^{(k)}) \le & \frac{(2^i-1) \xi}{2}\|  \mathbf{d}^{(k+1)}-\mathbf{d}^{(k)}\|^2
  + \nabla_{\dd} \mathcal{B}^*(\mathbf{d}^{(k)})(\mathbf{d}^{(k+1)}-\mathbf{d}^{(k)})\nonumber\\
  &+  \alpha_2 \sum_{t=1}^{T}  \left(\mathcal{R}( \mathbf{G}^*\dd_t^{(k)}  ) - \mathcal{R}( \mathbf{G}^*\dd_t^{(k+1)}  )\right).
\end{align}
%
%
%
As mentioned at the beginning of the section, the cost function $\mathcal{J}(\epsi,  \mathbf{d},\mathbf{c})$  is piece-wise polynomial and therefore satisfies the ``Kurdyca-Lojasiewicz''  property. Hence, the sequence defined by \eqref{eq:Puy1}, \eqref{eq:Puy2} with factor selection \eqref{eq:stepfactor} is either unbounded or converges to a critical point of $\mathcal{J}(\epsi,  \mathbf{d},\mathbf{c})$. Finally, let us note that the boundedness of $\{(\epsi^{(k)}, \mathbf{d}^{(k)},\mathbf{c}^{(k)})\}_k$ is usually observed in practice or is easy to enforce by adding box constraints to the optimization problem. \\

\section{Experiments}\label{sec:4}

In this section, we provide an experimental validation of the SR procedure proposed in section \ref{sec:3.3}. We focus on the problem of recovering a sequence of HR natural  images from blurry and LR observations. In section \ref{sec:4.1}, we provide a precise definition of the model parameters used to run our algorithms. In section \ref{sec:4problems}, we describe several algorithms of the state of the art which will serve as points of comparison with the proposed approach. In sections \ref{sec:4.2} and \ref{sec:4.3}, we respectively describe the databases and the figures of merit which will be used in our experiments. Finally, a discussion of the performance of the proposed SR methodology is provided in section \ref{sec:4.4}. \\

\subsection{Specification of the Model and Algorithm Parameters}\label{sec:4.1}
We first discuss the choice of the parameters appearing in the model described in section \ref{sec:2}.  
 In particular, we  specify the definitions of  $\mathcal{H}$, $\mathcal{P}$, $\mathbf{D}$,    $\mathbf{G}$ and $\mathcal{R}$.  We then  provide some details about the parameters used in our algorithm. 

The observation model $\mathcal{H}$ is defined as the composition of a  low-pass filtering and down-sampling operation. The low-pass filter is assumed to model the blurring effect induced by the camera transfer function.  In our simulations, we use an approximation of a Gaussian kernel  with a standard deviation equal to $1.12$, as proposed in \cite{Burt81}.   A down-sampling factor equal to  2 is considered.

The operator $\mathcal{P}$ is supposed to model a ``displaced frame difference'' (DFD):  $\mathcal{P}$ is thus defined as in \eqref{eq:Pdx} with the interpolation functions $\{\psi_i\}_{i=1}^n$ equal to bi-dimensional cubic cardinal  splines \cite{Unser91}.  
 This representation offers a reasonable  accuracy with a  complexity scaling linearly with the image dimension, see  appendix \ref{sec:app2} for further details. 
%
%

 The dictionary $\mathbf{D}$ is chosen so that natural images have a sparse representation as a combination of a few of its columns. 
 Several choices of such dictionaries have been proposed in the literature, see, \eg, \cite{Mallat2008Wavelet,Peleg14}.
 Hereafter, we consider a dictionary made up of discrete real-valued curvelets \cite{Candes02}; curvelets are known to yield sparse representations of piece-wise smooth functions.  
  The choice of a curvelet dictionary is also motivated by the existence of fast algorithms for the computation of the product between $\mathbf{D}$ and some vector, see \cite{Candes06}: this transform is based on a fast Fourier transform  and its complexity\footnote{As mentioned earlier, a linear complexity can be preserved by using, for example, a wavelet basis instead of a curvelet frame.} scales as $\mathcal{O}(n\log n)$. 
  
   Matrix $\mathbf{H}$ appearing in the cost-to-move function in \eqref{eq:cost2move} is chosen to be a Haar wavelet basis\footnote{We note  that evaluation of products $\mathbf{H}$ or $\mathbf{H}^*$  only requires  a linear complexity since they can be implemented by  fast wavelet transforms \cite[Chapter 7]{Mallat2008Wavelet}.}. In practice, we did not observe a significant difference in our results by using other types of wavelets; we thus essentially consider Haar wavelets for simplicity purposes.
 
To complete our discussion, let us elaborate on the choice of $\mathbf{G}$ and $\mathcal{R}$, characterizing the regularization imposed on the displacement field $\mathbf{d}_t$. In our simulations, we wish to enforce  either a  global or a piecewise regularity of the motion. We proceed as follows. The spatial derivatives of the motion are approximated by a ``finite difference'' scheme: each finite difference corresponds to a particular element of the matrix-vector product $\mathbf{G}^*\mathbf{d}_t$ (matrix $\mathbf{G}$ thus contains ``$\pm1$'' elements located at proper positions). The  regularity of the motion field is then enforced by constraining the function $\mathcal{R}(\mathbf{G}^*\mathbf{d}_t)$ to be small. In our experimentations, we choose $\mathcal{R}$ to be defined as in \eqref{eq:DefR} with a weighting vector $\mathbf{w}$ as in \cite{Xu12}. Further details are provided in appendix~\ref{sec:app2}. 
 
Besides, we notice that, although we have presented our SR procedure in the case of a mono-channel image-sequence observation in section \ref{sec:3}, its extension to a multi-channel setting (\eg, when 3-channel color images are available) is  straightforward and will be considered in our simulations. 

We now specify the choice of the algorithm parameters. As exposed earlier, we rely on the recursion \eqref{eq:Puy1}-\eqref{eq:Puy2} described in section \ref{sec:3.3} to search for a critical point of the cost function in \eqref{eq:prob2} with $p=1$.
  Each step of the recursion \eqref{eq:Puy1}-\eqref{eq:Puy2} is solved via an ADMM procedure. Details on the ADMM steps are given  in appendix \ref{app:attouch}. The ADMM solvers   involve  minimizations  by a gradient descent procedure. In our implementation, we choose a quasi-Newton descent method adapted to our high-dimensional problem, namely a limited-memory Broyden-Fletcher-Goldfarb-Shanno (L-BFGS) procedure with a line-search routine  based on the strong Wolf conditions \cite{Nocedal99}.
  We stop the ADMM recursions after 20 iterations and the global 2-step recursions after 20 iterations, since we observed no significant improvements of the results for a larger number of iterations.
  
The super-resolved images $\xx_t$'s  are initialized  by  Lanczos interpolation of their low-resolution counterparts. Motion fields  are initialized with  an  upscaled  optic-flow estimate obtained by applying algorithm  \cite{Xu12} on the low-resolution observations.  To perform a fair comparison with  the ``multi-frame'' SR algorithm of {\it Mitzel et al.}  \cite{Mitzel09} described in the next section,  we also ran our algorithm with an initial motion field computed with the optic-flow algorithm  \cite{Zach07}}. {For both initializations, t}he upscaling from the low-resolution optic-flow estimate to the high-resolution motion field is done with a Lanczos interpolation.  The values of the other parameters of our algorithm are given in table \ref{tab:1}. These parameters have been tuned experimentally  to lead to a reasonable tradeoff between visual inspection and error measurements for the data-set benchmark presented in section \ref{sec:4.2}. \\


\begin{table}
\begin{center}
\small
\hspace{-0cm}\begin{tabular}{|c|c|c|c|c|c|c|c|c|}
\hline
  $\alpha_1 $& $\rho_1$&$\alpha_2$&$\rho_2$&$\alpha_3$ &$\rho_3$& $\gamma$&$\rho$&$\alpha^{(0)}$\\
\hline
	5e-1&1e2&8e3&1e1&1e1 &1e-2	&1e0&  1e0   &2e2   \\
\hline
\end{tabular}\vspace{0.5cm}
\caption{Algorithm  parameter setting. Parameters $\alpha_1$, $\alpha_2$ and $\alpha_3$ appear in the cost function \eqref{eq:prob2}. Parameters $\gamma$ and $\alpha^{(0)}$ specify  the 2 steps  \eqref{eq:Puy1} and  \eqref{eq:Puy2}.   Parameters $\rho_1$, $\rho_2$, $\rho_3$ and $\rho$ are auxiliary factors used in the ADMM recursions.     \label{tab:1}}
\end{center}
\end{table}

 \subsection{Algorithm Benchmark}\label{sec:4problems}

 The assessment  of the proposed algorithm relies on a  comparison with a benchmark of three state-of-the-art methods:\\
 
\begin{itemize}
\item[\textbullet] the ``single-frame'' SR algorithm of {\it Peleg et al., 2014}  \cite{Peleg14},  
\item[\textbullet]  the ``kernel-regression'' SR algorithm of  {\it Takeda et al., 2009}  \cite{Takeda09,Milanfar10},
\item[\textbullet] the ``multi-frame'' SR algorithm of {\it Mitzel et al., 2009}  \cite{Mitzel09}. \\
  \end{itemize}
  
 These algorithms are adapted to the super-resolution of videos exhibiting non-homogeneous displacements.   Moreover,  each of these three algorithms is a state-of-the-art method representing  a class of SR algorithms. The algorithm of {\it Peleg et al., 2014}  implements a ``single-frame'' SR method based on a statistical learning procedure with sparse representations; the  algorithm by {\it Takeda et al., 2009}  is a SR  method based on a ``multidimensional kernel regression'' fitting   the low-resolution observations;  the algorithm of  {\it Mitzel et al., 2009}   implements a multi-frame SR method using  a  quadratic relaxation scheme for high-accuracy optic-flow estimation \cite{Zach07}. Finally, we also compare the performance obtained with the proposed method with two standard  spatial interpolation techniques, namely\\  
\begin{itemize}
\item[\textbullet]  the basic {\it nearest neighbor} upscaling (block interpolation),
\item[\textbullet]   {\it  Lanczos} interpolation \cite{Turkowski90}.\\
  \end{itemize}
Note that in order to treat color image sequences, algorithms   only supporting  gray-level  images  are run independently on the three spectral bands. \\

\subsection{Data-set Benchmark}\label{sec:4.2}
We evaluate the performance of the algorithms using a benchmark of three image sequences:\\
\begin{itemize}
\item[\textbullet]   A synthetic sequence  from the  \textit{MPI Sintel} data set \cite{Sintel}. This recent data set, which is derived from the open source 3D animated short film,  was originally created for the evaluation of optical flows. The synthesized image sequences are realistic and particularly challenging: on the one hand, displacement  fields are characterized by large amplitudes, discontinuities,  blur  or defocus effects; on the other hand,  the image sequence presents many occlusions, specular reflections  or  atmospheric effects. In our simulations, we focus on a region of interest  of $436 \times 512$ pixels and on the 8 first images.  The first and last images of the  ``\textit{bandage}''  data-set sequence are displayed in Fig. \ref{figNew0}. In the following, we will refer to this sequence as data set \#1.
\item[\textbullet] A real sample of the standard ``\textit{foreman}''  video\footnote{Image sequences are part of the \textit{Derf Collection}, which can be downloaded at {\it https://media.xiph.org/video/derf} [Online; posted 15-October-2015]}. In our simulations, we focus on a region of interest  of $256 \times 256$ pixels and on the 10 first images.   The first and last images of this data set are displayed in Fig. \ref{figNew1}. In the following, we will refer to this sequence as data set \#2.
\item[\textbullet]  A real sample of the challenging  ``\textit{football}''  video$^{12}$, which  exhibits non-homogeneous and large displacements, as well as  multiple occlusions.  In our simulations, we focus on a region of interest  of $256 \times 256$ pixels and on the 10 first images.  The first and last images of this  data set  are displayed in Fig. \ref{figNew2}. In the following, we will refer to this sequence as data set \#3.\\

  \end{itemize}
  The    images of these sequences are composed of three spectral bands, each one is coded in 8 bits. We 
  create the LR images by applying function $\mathcal{H}$ on these sequences. This function first filters the discrete signal by a Gaussian kernel  of standard deviation equal to 1.12 and then  down-samples the result by a factor of  2, see \cite{Burt81}. \\

\subsection{Evaluation Procedure}\label{sec:4.3}

The performance of the algorithms is assessed   in terms of  reconstruction of the super-resolved image and estimation of the motion  field. We describe the figures of merit used in our assessments hereafter. Let  $\{ \hat \xx_t\}_{t=0}^T$  (resp.  $\{ \hat{\dd}_t\}_{t=1}^T$) denote the estimated image sequence (resp. displacements) and  $\{\xx_t^{true}\}_{t=0}^T$ (resp.  $\{ \dd_t^{true}\}_{t=1}^T$) the corresponding ground truth.
 Standard criteria \cite{Nasrollahi2014Superresolution} to measure the image sequence reconstruction accuracy  are the peak-to-signal ratio (PSNR) at time $t$: 
\begin{align*}
\text{PSNR}(t)= 20 \log_{10}\frac{n\| \xx_t^{true}\|_{\infty}}{  \| \xx_t^{true} -  \hat \xx_t  \|_2 },
\end{align*}
and the correlation coefficient (CC) at time $t$:
\begin{align*}
\text{CC}(t)=     \frac{  (\xx_t^{true} - \mu_{\xx_t^{true}})^* ( \hat \xx_t - \mu_{\hat \xx_t})}{\| \xx_t^{true} - \mu_{\xx_t^{true}}\|_2\| \hat \xx_t - \mu_{\hat \xx_t}\|_2 },
\end{align*}
where we have denoted the arithmetic mean of vector  $\hat \xx_t$ and $\xx_t^{true}$  by  $\mu_{\hat \xx_t}$ and $\mu_{\xx_t^{true}}$.  We evaluate the accuracy of the estimated motion fields with the time-averaged  Mean End Point Error (MEPE): 
\begin{align*}\text{MEPE}=  \frac{1}{nT}\sum_{t=1}^T  \| \dd_t^{true} -  \hat \dd_t \|_1 ,
\end{align*}
%
%
and the time-averaged Mean Barron Angular Error (MBAE) in degrees \cite{Baker2011Database}: 

\begin{align*}
\text{MBAE} = \frac{1}{nT} \sum_{t=1}^T\sum_{s=1}^n  \text{arcos}\left(\frac{ 1+\dd_t^{true}(s)   \hat \dd_t (s)+\dd_t^{true}(s+n)   \hat \dd_t (s+n)}{ \sqrt{(1+\hat \dd_t(s)^2+\hat \dd_t(s+n)^2)\;(1+ \dd^{true}_t(s)^2+ \dd^{true}_t(s+n)^2)}}\right),
\end{align*}
where we have adopted the convention that the two $n$-dimensional components of motion have been sorted  one after the other in vectors $\hat \dd_t$ and $\dd_t^{true}$.

In order to compare   the different algorithms (algorithm  \cite{Takeda09} does not support  large images and exclude pixels at the image  border), the criteria PSNR and  CC  are evaluated on a spatial window of size of $240\times240$ cropped in the image sequences.   \\

\subsection{Results and  Discussion}\label{sec:4.4}

\begin{table}[!t]
\begin{center}
\small
\hspace{0cm}\begin{tabular}{|c|c|c|c|c|c|c|}
\hline
\multicolumn{1}{|c|}{} & \multicolumn{3}{|c|}{ PSNR($t$) } & \multicolumn{3}{|c|}{ CC($t$)}   \\
\cline{2-7}
&set  \# 1 &set  \# 2&set \# 3&set \# 1 & set \# 2& set \# 3 \\
\cline{2-7}
&  $t$=5 &$t$=7&$t$=7&$t$=5 & $t$=7& $t$=7\\
\hline
Nearest neighbor&27.236&22.639	 &	22.656	&0.9771  & 0.9648&0.9556\\
Lanczos interpolation \cite{Turkowski90} &27.845&24.571	 &23.040&	0.9800	&0.9775&0.9579 \\
Single-frame SR  \cite{Peleg14}  &28.359&32.518	&	23.971	&0.9815&0.9949& 0.9684\\
Kernel-regression SR \cite{Takeda09,Milanfar10} &28.944&33.275	&23.394	&0.9838&0.9957&0.9643  \\
 Multi-frame SR \cite{Mitzel09} &29.935&32.295	 &21.948	&0.9844&0.9939& 0.9491 \\
 Proposed (optic-flow init. \cite{Zach07}) &30.634&\textbf{35.027}	 &25.007	&0.9868&\textbf{0.9969}& 0.9750 \\
 Proposed (optic-flow init. \cite{Xu12}) &\textbf{30.790}&	{34.305} &\textbf{25.302}	&\textbf{0.9872}&{0.9963}&\textbf{0.9767}\\
\hline
\end{tabular}\vspace{0.5cm}
\caption{Accuracy of super-resolved image estimates  in terms of PSNR and CC at time $t$. 
 \label{Newtab:1}}\
\end{center}
\end{table}

Table \ref{Newtab:1}  presents the accuracy of the different algorithms  in terms of PSNR and CC.   We evaluated these criteria at $t=5$ for data set \#1 and at  $t=7$ for data sets \#2 and \#3. 
 We first note that our SR method yields better figures of merit than the other methods  for the different data sets of the benchmark. It improves slightly the CC and substantially the PSNR (more than a unit) for each data-set configuration.   
Second, the estimates released by the proposed approach seem to achieve a good quality level irrespective of the considered data set:
 on the contrary, the multi-frame SR algorithm \cite{Mitzel09} performs fairly well on data set \#1 but its performance collapses on data set \#3; 
  the kernel-regression  SR algorithm \cite{Takeda09,Milanfar10} obtains good results for data set \#2 while  yielding only a slight  increase of the accuracy with respect to a Lanczos interpolation on data set \#3; 
  the single-frame SR algorithm \cite{Peleg14} has a good behavior for data set \#3  but is less competitive for data set \#1. {Third,  the performance of our algorithm seems to be comparable for different  motion initializations,  in particular for initial motion fields obtained from the  optic-flow algorithms   of \cite{Xu12} or \cite{Zach07}.}

 The improvement brought by the proposed method can also be seen by a visual inspection of the reconstructed images in 
  Fig.~\ref{figNew3},  \ref{figNew4}, \ref{figNew6} and \ref{figNew7}. 
 We can first underline the enhancement provided by the inclusion of some motion information in the SR reconstruction process {by comparing the estimates released by  the ``single-frame'' and the ``multi-frame''/``sequential'' algorithms}. 
  {In Fig.~\ref{figNew3} and \ref{figNew6}}, one can notice that the estimated contours and the texture are over-smoothed if no motion information is included. This is for example visible by inspecting  the fuzzy girl's eyebrow or the smoothed scales of the little dragon in Fig.~\ref{figNew3},  distinguishing the tongue of the foreman or analyzing the texture of the grass field of the football game in Fig.~\ref{figNew6}. In comparison, our algorithm enhances  the reconstruction accuracy of these details as visible in Fig.~\ref{figNew4} and \ref{figNew7}. The drawback of including motion is that, as it can be noticed for the little dragon,  errors in  motion discontinuity estimation  may  induce imprecision on the contours and  lead to some undesirable oscillations. 

 Although not as accurate as the proposed method, we note  the good performance of the  single-frame SR algorithm proposed in \cite{Peleg14}. Clearly,  it is  competitive with  other state-of-the-art approaches exploiting motion information. This is probably due to the relevance of  the sparse prior employed by the  single-frame SR algorithm \cite{Peleg14}. This is particularly striking when the motion in the video is too difficult to exploit  by the multi-frame or kernel-regression SR algorithms, as  shown for the challenging football sequence in Fig.~\ref{figNew6} and \ref{figNew7}. Let us also mention that results obtained with a kernel-regression SR strategy reveal a slight enhancement in comparison to standard spatial interpolation techniques, which is probably induced by the implicit introduction of the motion information \textit{via} the modeling of the local spatio-temporal structures of the sequence.

 
Our experiments  also emphasize several examples where a sequential SR setup can solve some reconstruction ambiguities which can be difficult to treat in a multi-frame framework.  
 In Fig.~\ref{figNew4} and \ref{figNew7},  some erroneous reconstructions, which do not appear in the proposed method, can be noticed in the multi-frame estimates: 
  for example,  {artefacts in the girl's eye  in Fig.~\ref{figNew4}, deformations of the foreman's tongue  and the fuzziness of the stripes of the football player's trousers in Fig.~\ref{figNew7}}. 
  Indeed, matching all the images of the sequence with  a reference frame is often a more difficult task than estimating motions between consecutive frames. 
  In the former situation, motion estimation has to deal  with large displacements between distant frames whereas, in the latter setup, the problem simplifies into the estimation of a succession of small displacements. 
  In other words,  a SR multi-frame setup will try to match images of the sequence which could apparently seem independent, with the potential drawback of estimating erroneous structures. 
  On the other hand, a SR sequential setup  propagates information through  consecutive frames and may better succeed in modeling  the overall dependences in the image sequence. 
  {One could nevertheless argue that the estimation of inter-frame motions could also lead to error propagation if the motion estimates are inaccurate. 
  This is not what we observed in our simulations: motion errors are usually absorbed by the error terms $\boldsymbol{\epsilon}_t$ (which increase in the region where the  motion is badly estimated)}. 
  This is illustrated in Fig.~\ref{figNew8}: we observe that {$\boldsymbol{\epsilon}_t$ may be large on the contours of the characters (where the quality of the motion estimation is typically low)} but the PSNR is nevertheless stable across the reconstructed image sequence. 
  Therefore, as observed from our simulations, a sequential SR approach is usually better conditioned to  deal with videos such as data sets \#1 and \#3, which exhibit large displacements and/or occlusions.

\begin{table}[t]
\begin{center}
\small
\hspace{0cm}\begin{tabular}{|c|c|c|c|c|}
\hline
\multicolumn{1}{|c|}{} &\multicolumn{2}{|c|}{LR estimate} &\multicolumn{2}{|c|}{SR estimate}   \\
\cline{2-5}
 &{ MEPE} & { MBAE}&{ MEPE} & { MBAE}   \\
 \hline 
\textit{Zach et al., 2007} \cite{Zach07}. &1.319 &  24.988  & \textbf{1.302} & \textbf{ 24.975}\\
\textit{Xu et al., 2012} \cite{Xu12}. &1.342& 25.592 &\textbf{1.320}&\textbf{ 25.545}\\
\hline
\end{tabular}\vspace{0.5cm}
\caption{Accuracy of low-resolved  or super-resolved optic-flow estimates  in terms of MEPE and MBAE, with respect to the motion initialization algorithm. 
 \label{Newtab:2}}\
\end{center}
\end{table}


Finally, let us notice that there is a positive interaction between the estimation of the motion fields and the HR images:   
intuitively, it is clear that a good estimation of the HR image sequence will improve the quality of the estimated motion fields; similarly, a good estimation of the super-resolved motion fields will enhance the accuracy of the estimated image sequence. Although this positive interaction is difficult to ensure from a theoretical side, we have often observed it in practice.  
{We illustrate in Table \ref{Newtab:2} and Fig.~\ref{figNew5} the benefit of refining the motion estimation through our iterative procedure for the synthetic data set \#1, independently of the initial motion estimate. In Table \ref{Newtab:2}, we can notice a slight gain in terms of MBAE and MEPE in comparison to a direct estimation of the motion from the LR observations with the methods presented in \cite{Xu12} or \cite{Zach07} (which serve as initializations for our algorithm, see section \ref{sec:4.1}).  {More interestingly,} we note {in Fig.~\ref{figNew5}} that the motion field {released by} the proposed approach exhibits sharper discontinuities than those output by \cite{Xu12} or \cite{Zach07}. \\

\section{Conclusion}

 We have presented a {new} methodology to solve video SR problems,  \ie,  to reconstruct a HR image sequence from LR observations. 
 The HR sequence is entirely described by a parametric non-linear sequential model{, which connects the different images of the sequence. It is parametrized by a final condition, a sequence of non-global displacement fields and a sequence of additive noises.}
  {In order to compensate for the ill-posedness of the video SR problem, we considered priors enforcing some forms of sparsity on the unknown parameters of the system.} 
  The joint estimation of the final condition, the displacement and the noise sequences was expressed as a constrained minimization problem which, in the   general case,  is high-dimensional, non-differentiable and non-convex.  
  We provided  elementary building blocks to tackle each of these difficulties, and, by gathering them,  designed a convergent optimization algorithm enjoying a  complexity (per iteration) linear in the problem dimensions. 
%
Our numerical simulations on several video benchmarks show that the proposed SR method {is competitive with}  state-of the-art. In particular, the gain appears to be particularly  important for  videos involving complex motions with  large amplitudes and occlusions.\\
 

   \section*{Acknowledgements}
The authors wish to acknowledge C. Deltel and S. Campion for their technical support in numerical simulations.\\


\appendix

\section{  Proof of \eqref{eq:grad1}-\eqref{eq:foward1}}\label{sec:app1}
The proof of this specific \textit{backward} optimal control solution follows the sketch of the demonstration  for the more standard \textit{forward} problem presented in \cite{Bertsekas99}.   We will focus on the following optimization problem: 
\begin{align}\label{eq:optimP}
\argmin_{(\epsi,\dd,\cv)}   \mathcal{T}(\xx=\mathcal{Q}(\epsi,\dd,\mathbf{c}),\epsi,\dd,\cv),
\end{align}
where $\mathcal{T}$ denotes some objective function to be defined below. We recall that,  given ($\epsi,\dd$, $\mathbf{c})$,  the function $\mathcal{Q}(\epsi,\dd,\mathbf{c})$  determines  a unique vector $\xx=\mathcal{Q}(\epsi,\dd,\mathbf{c})$ satisfying the constraints in \eqref{eq:prob2}, see section \ref{sec:3.1}.
In this appendix, we will use the following short-hand notation for the constraints in \eqref{eq:prob2}:
\begin{equation}\label{eq:constPart1}
  \left\{\begin{aligned}
&\xx_t=\mathcal{F}_t(\xx_{t+1}, \epsi_{t+1},\dd_{t+1}),&\hspace{0.2cm} 0\leq t \leq  T-1, \\
&\xx_T=\mathbf{D} \cv,&\hspace{0.2cm}
  \end{aligned}\right.
\end{equation}
with $\mathcal{F}_t(\xx_{t+1}, \epsi_{t+1},\dd_{t+1})\triangleq \mathcal{P}(\xx_{t+1},\dd_{t+1})+\boldsymbol{\epsilon}_{t+1}$.
We will also alleviate the notation for the constraint $\xx=\mathcal{Q}(\epsi,\dd,\mathbf{c})$ by denoting this vector simply by $\xx$. Therefore, $\xx$ should  be understood as a function of $\epsi,\dd,\mathbf{c}$ and no longer as an independent variable.

The proof of \eqref{eq:grad1}-\eqref{eq:foward1} is made of two different parts, in which  we study  different instances of  optimization problem \eqref{eq:optimP}. 
 In a first step, we will consider an objective function only depending on the initial state\footnote{As mentioned previously, $\xx_0$ must be understood as a function of $\epsi,\dd,\mathbf{c}$.}: 
\begin{align}\label{eq:optim1}
\mathcal{T}(\xx,\epsi,\dd,\cv)\triangleq \mathcal{G}_0(\xx_0).
\end{align}
 Then, in a second step, we will  come back to the more general problem \eqref{eq:unprob}, \ie,  an optimization problem where the objective function $\mathcal{T}$ will match the cost function, $\mathcal{J}$, given in  \eqref{eq:costfunctionun}:  
 \begin{align}\label{eq:optim2}
 \mathcal{T}(\xx,\epsi,\dd,\cv)\triangleq  \mathcal{G}_0(\xx_0)+\sum_{t={1}}^{T-1} \mathcal{G}_t(\xx_t,\epsi_t,\dd_t) + \mathcal{G}_T(\xx_T,\epsi_T,\dd_T,\cv)\triangleq\mathcal{J}(\xx,\epsi,\dd,\cv).\\
 \nonumber
 \end{align}

\textbf{First Part of the Proof.}\, We begin by considering problem \eqref{eq:optimP}  with the objective function   \eqref{eq:optim1}.
 By the chain rule of derivation applied to \eqref{eq:constPart1} at some point in the set 
 \begin{align}\label{eq:setCApp}
\left\{ (\xx',\epsi', \dd',\mathbf{c}') \left| 
\begin{aligned}
\xx'_t=\mathcal{F}_t(\xx'_{t+1}, \epsi'_{t+1},\dd'_{t+1}),&\hspace{0.2cm} 0\leq t \leq  T-1 \\
\;\;\xx'_T=\mathbf{D} \cv'&\hspace{0.2cm}
\end{aligned}
\right.
\right\},
\end{align}
  we can decompose the gradients  into the products: 
\begin{equation}\label{eq:system1}
\left\{\begin{aligned}
\nabla_{\epsi_t} \mathcal{T}(\xx',\epsi',\dd',\cv')&=\nabla_{\epsi_t} \mathcal{F}_{t-1}^*( \xx_t',\epsi_t',\dd_t') \, \nabla_{\xx_{t-1}}\mathcal{F}_{t-2}^* \, \cdots\nabla_{\xx_2}\mathcal{F}_{1}^* \,  \nabla_{\xx_1}\mathcal{F}_{0}^* \,  \nabla_{\xx_0} \mathcal{G}_0(\xx_0'),\\
\nabla_{\dd_t} \mathcal{T}(\xx',\epsi',\dd',\cv')&=\nabla_{\dd_t} \mathcal{F}_{t-1}^*( \xx_t',\epsi_t',\dd_t') \, \nabla_{\xx_{t-1}}\mathcal{F}_{t-2}^* \, \cdots\nabla_{\xx_2}\mathcal{F}_{1}^* \,  \nabla_{\xx_1}\mathcal{F}_{0}^* \,  \nabla_{\xx_0} \mathcal{G}_0(\xx_0'),\\
\nabla_{\cv} \mathcal{T}(\xx',\epsi',\dd',\cv')&=\mathbf{D}^*\nabla_{\xx_T}\mathcal{F}_{T-1}^* \,  \cdots  \nabla_{\xx_2}\mathcal{F}_{1}^* \,  \nabla_{\xx_1}\mathcal{F}_{0}^* \,  \nabla_{\xx_0} \mathcal{G}_0(\xx_0'),
\end{aligned}\right.
\end{equation}
 where  we recall that $\nabla_{\xx_t} \mathcal{F}_{t-1} $ denotes the Jacobian matrice of $\mathcal{F}_{t-1}$ with respect to function $\xx_t$  evaluated at $(\xx_t',\epsi_t',\dd_t')$ and $\nabla_{\xx_t} \mathcal{F}_{t-1} ^*$ its transpose.  
  We can rewrite gradients in \eqref{eq:system1} in order to exhibit their recursive structures.  By defining the forward recursion 
\begin{equation}\label{eq:fwrd1}
\left\{\begin{aligned}
&\zv_0=\nabla_{\xx_0} \mathcal{G}_0(\xx_0'),\\
&\zv_t=\nabla_{\xx_t}\mathcal{F}_{t-1}^*   \zv_{t-1},\hspace{0.3cm}1\leq t \leq  T,
\end{aligned}\right.
\end{equation}
we obtain the following rewriting:
\begin{equation}\label{eq:gradH1}
\left\{\begin{aligned}
&\nabla_{\boldsymbol{\epsilon}_{t}} \mathcal{T}(\xx',\epsi',\dd',\cv')=\nabla_{\boldsymbol{\epsilon}_{t}} \mathcal{F}_{t-1}^*(\xx_t',\epsi_t',\dd_t') \zv_{t-1},\hspace{0.3cm}1\leq t \leq  T,\\
&\nabla_{\dd_{t}} \mathcal{T}(\xx',\epsi',\dd',\cv')=\nabla_{\dd_{t}} \mathcal{F}_{t-1}^*(\xx_t',\epsi_t',\dd_t') \zv_{t-1},\hspace{0.3cm}1\leq t \leq  T,\\
&\nabla_{\cv} \mathcal{T}(\xx',\epsi',\dd',\cv')= \mathbf{D}^* \zv_T.\\
\end{aligned}\right.\\
\end{equation}\vspace{0.5cm}

\textbf{Second Part of the Proof.}\,  We now consider  problem \eqref{eq:optimP} with objective function \eqref{eq:optim2}. 
By making a change of variables, we want to obtain a rewriting of function \eqref{eq:optim2}  with a structure  analogous to   \eqref{eq:optim1}, so that  the gradients are given by a recursion of the form \eqref{eq:fwrd1}-\eqref{eq:gradH1}.  In other words, by making some change of variables  we intend to rewrite the sum of functions in \eqref{eq:optim2} as a unique function depending solely on an ``initial state''. In order to do so, let us define  variables $\boldsymbol{\kappa}_i$'s recursively as follows:   
\begin{equation*}
\left\{\begin{aligned}
&\boldsymbol{\kappa}_T=0,\\
&\boldsymbol{\kappa}_{T-1}(\xx_T,\epsi_T,\dd_T,\cv)=\boldsymbol{\kappa}_{T}+\mathcal{G}_{T}(\xx_T,\epsi_T,\dd_T,\cv),\\
&\boldsymbol{\kappa}_{t-1}(\xx_t,\epsi_t,\dd_t,\cv)=\boldsymbol{\kappa}_{t}+\mathcal{G}_{t}(\xx_{t},\epsi_{t},\dd_t),\hspace{0.3cm}T-1\geq t \geq  1.
\end{aligned}\right.
\end{equation*}
We then obtain that  
$$\boldsymbol{\kappa}_0(\xx,\epsi,\dd,\cv)=\sum_{t={1}}^{T-1} \mathcal{G}_t(\xx_t,\epsi_t,\dd_t) + \mathcal{G}_T(\xx_T,\epsi_T,\dd_T,\cv),$$ 
 and  
  the objective function $\mathcal{T}$  given in \eqref{eq:optim2} can be rewritten as 
\begin{align}\label{eq:ObjStep2_}
\mathcal{T}(\xx,\epsi,\dd,\cv)=\boldsymbol{\kappa}_0(\xx,\epsi,\dd,\cv)+\mathcal{G}_0(\xx_0).
\end{align}
Considering the following change of variables $\tilde \xx_t \triangleq \begin{pmatrix}\xx_t \\ \boldsymbol{\kappa}_t \end{pmatrix}$, we then have that the right-hand side of \eqref{eq:ObjStep2_} can be rewritten as a function of $\tilde{\xx}_0$ only. In the sequel, we will use the following specific notation to emphasize this fact: 
\begin{align}\label{eq:defG0tilde}
\mathcal{ \tilde G}_0(\tilde \xx_0) = \boldsymbol{\kappa}_0 + \mathcal{G}_0(\xx_0).
\end{align}
Moreover, it is easy to see that functions $\tilde \xx_t$'s satisfy  the following backward recursion:  
\begin{equation}\label{eq:backwardconst}
\left\{\begin{aligned}
&\tilde \xx_t= \mathcal{\tilde F}_t(\tilde \xx_{t+1},\epsi_{t+1},\dd_{t+1}) ,\hspace{0.3cm}T-1\geq t \geq  0,\\
&\tilde \xx_T=\mathcal{\tilde F}_T(\epsi_T,\dd_T,\cv),\\
\end{aligned}\right.
\end{equation}
where
\begin{align}
\mathcal{\tilde F}_t(\tilde \xx_{t+1},\epsi_{t+1},\dd_{t+1})&=\begin{pmatrix} \mathcal{F}_t(\xx_{t+1},\epsi_{t+1},\dd_{t+1})\\ \boldsymbol{\kappa}_{t+1}+\mathcal{G}_{t+1}(\xx_{t+1},\epsi_{t+1},\dd_{t+1})\nonumber \end{pmatrix},\\
\tilde{\mathcal{F}}_T(\epsi_T,\dd_T,\cv) &=\begin{pmatrix}\mathbf{D}\cv \\\mathcal{G}_T(\mathbf{D}\cv,\epsi_T,\dd_T,\cv)\nonumber\end{pmatrix}.
\end{align}
   We remark that the cost function \eqref{eq:defG0tilde}, recursion \eqref{eq:backwardconst} and the set
  \begin{align}\label{eq:setCApp2Part}
\left\{ (\tilde \xx',\epsi', \dd',\mathbf{c}') \left| 
\begin{aligned}
\tilde \xx'_t=\mathcal{\tilde F}_t(\tilde\xx'_{t+1}, \epsi'_{t+1},\dd'_{t+1}),&\hspace{0.2cm} 0\leq t \leq  T-1 \\
\;\; \tilde \xx'_T=\mathcal{\tilde F}_T(\epsi_T',\dd_T',\cv') &\hspace{0.2cm}
\end{aligned}
\right.
\right\},
\end{align}
    have respectively the same structure as  \eqref{eq:optim1}, \eqref{eq:constPart1} and \eqref{eq:setCApp}. We can then apply the  result obtained previously and get the gradients of $\mathcal{T}$ 
    using the same reasoning as the one made to derive \eqref{eq:fwrd1}-\eqref{eq:gradH1}. More specifically,  let $(\tilde \xx',\epsi',\dd',\cv')$ be some point  in  \eqref{eq:setCApp2Part}, and let $\tilde \zv_t$ be an ``adjoint'' variable verifying: 
\begin{equation}\label{eq:recInt}
\left\{\begin{aligned}
&\tilde \zv_0=\nabla_{\tilde \xx_0} \mathcal{ \tilde G}_0(\tilde \xx_0'),\\
&\tilde \zv_t= \nabla_{\tilde \xx_{t}} \mathcal{\tilde F}_{t-1}^* \tilde \zv_{t-1},\hspace{0.3cm} 1\leq t \leq  T,
\end{aligned}\right.
\end{equation}
where the Jacobian matrix of  $\mathcal{\tilde F}_{t-1}$  evaluated at  some point $(\tilde \xx_t',\epsi_t',\dd_t')$ is denoted $\nabla_{\tilde \xx_{t}} \mathcal{\tilde F}_{t-1} $. 
Using \eqref{eq:gradH1}, we obtain the  following expressions:
\begin{equation*}
\left\{\begin{aligned}
&\nabla_{\dd_{t}} \mathcal{T}(\tilde \xx',\epsi',\dd',\cv')=\nabla_{\dd_{t}} \mathcal{\tilde F}_{t-1}^*(\tilde \xx_t',\epsi_t',\dd_t') \tilde \zv_{t-1},\hspace{0.3cm} 1\leq t \leq  T,\\
&\nabla_{\boldsymbol{\epsilon}_{t}} \mathcal{T}(\tilde\xx',\epsi',\dd',\cv')=\nabla_{\boldsymbol{\epsilon}_{t}} \mathcal{\tilde F}_{t-1}^*(\tilde \xx_t',\epsi_t',\dd_t') \tilde \zv_{t-1},\hspace{0.3cm} 1\leq t \leq  T,\\
&\nabla_{\cv} \mathcal{T}(\tilde\xx',\epsi',\dd',\cv')=\nabla_{\cv} \mathcal{\tilde F}_{T}^*(\epsi_T',\dd_T',\cv') \tilde \zv_T.
\end{aligned}\right.
\end{equation*}

To finalize the proof, we re-express recursion \eqref{eq:recInt} by developing it with respect to the two different components of the adjoint variable  $\tilde \zv_t \triangleq \begin{pmatrix}\zv_t\\\omega_t\end{pmatrix}$, where the $\zv_t$'s have the dimension of $\xx_t$'s and $\omega_t$'s are scalars. Particularizing the first equation in \eqref{eq:recInt}  by taking \eqref{eq:defG0tilde} into account, we obtain  
\begin{align}\label{eq:SysytAdDev3}
\begin{pmatrix}\zv_0\\\omega_0\end{pmatrix}&=\begin{pmatrix} \nabla_{\xx_0} \mathcal{ \tilde G}_0(\tilde \xx_0') \\  \nabla_{\boldsymbol{\kappa}_0} \mathcal{ \tilde G}_0(\tilde \xx_0')\end{pmatrix}= \begin{pmatrix} \nabla \mathcal{G}_0(\xx_0') \\  1\end{pmatrix}.
\end{align}
Moreover, using the definition of $\tilde{\mathcal{F}}_t$, the second equation in \eqref{eq:recInt} leads to   
\begin{align}\label{eq:SysytAdDev1}
\begin{pmatrix} \zv_{t} \\ \omega_{t}\end{pmatrix} &= \begin{pmatrix}\nabla_{ \xx_{t}} \mathcal{ F}_{t-1}^*   &  \nabla_{\xx_{t}}  \mathcal{G}_{t}(\xx_{t}',\epsi_t',\dd_t')\\0 & 1\end{pmatrix}\begin{pmatrix} \zv_{t-1} \\ \omega_{t-1}\end{pmatrix},\quad 1\leq t \leq  T-1,\\
\label{eq:SysytAdDev2}
\begin{pmatrix} \zv_{T} \\ \omega_{T}\end{pmatrix} &= \begin{pmatrix}\nabla_{ \xx_{T}} \mathcal{ F}_{T-1}^*   &  \nabla_{\xx_{T}}  \mathcal{G}_{T}(\xx_T',\epsi_T',\dd_T',\cv')\\0 & 1\end{pmatrix}\begin{pmatrix} \zv_{T-1} \\ \omega_{T-1}\end{pmatrix} .
\end{align}

Equations \eqref{eq:SysytAdDev3}-\eqref{eq:SysytAdDev2} imply that $\omega_{t}=1$ $\forall t$; moreover, the recursion in $\zv_t$ is equivalent to \eqref{eq:foward1}.\\ 

\section{The Alternating Direction Method of Multipliers}\label{app:ADMM}
The alternating direction method of multipliers (ADMM) focusses on the following type of optimization problems:
\begin{align}\label{eq:ADMMstdproblem}
\min_{\zz_1\in\Zc_1,\zz_2\in\Zc_1} \mathcal{G}_1(\zz_1)+\mathcal{G}_2(\zz_2),\quad 
\mbox{ s.t.  $\Am\zz_1+\Bm \zz_2=\mathbf{0}_r$},
\end{align}
where 
 $\Am\in\Rb^{r\times n_1}$, $\Bm\in\Rb^{r\times n_2}$,  $\mathcal{G}_1:\Rb^{n_1}\rightarrow\Rb$, $\mathcal{G}_2:\Rb^{n_2}\rightarrow\Rb$ are closed, proper and convex functions, and $\Zc_1$, $\Zc_2$ are non-empty convex sets. We note that the conditions on $\mathcal{G}_1$ and $\mathcal{G}_2$ are pretty mild; in particular, $\mathcal{G}_1$ and $\mathcal{G}_2$ are not required to be differentiable and can take on infinite values. 

ADMM is an iterative procedure inspired by the well-known method of multipliers \cite{Bertsekas99}. It searches for a minimizer of \eqref{eq:ADMMstdproblem} by sequentially minimizing the corresponding augmented Lagrangian with respect to each primal variables $\zz_1$ and $\zz_2$, before updating a dual variable $\mathbf{u}\in\Rb^r$. Formally, the ADMM recursions take the form:
\begin{align}
\zz_1^{(k+1)}&=\argmin_{\zz_1\in\Zc_1} \mathcal{G}_1(\zz_1)+\frac{\rho}{2} \|\Am\zz_1+\Bm \zz_2^{(k)}+\uv^{(k)}\|^2_2,\label{eq:ADMMstdformA}\\[0.1cm]
\zz_2^{(k+1)}&=\argmin_{\zz_2\in\Zc_2} \mathcal{G}_2(\zz_2)+\frac{\rho}{2} \|\Am\zz_1^{(k+1)}+\Bm \zz_2+\uv^{(k)}\|^2_2,\label{eq:ADMMstdformB}\\[0.1cm]
\uv^{(k+1)}&=\uv^{(k)}+\Am\zz_1^{(k+1)}+\Bm\zz_2^{(k+1)},\label{eq:ADMMstdformC}
\end{align}
for some $\rho>0$. 

ADMM has recently sparked a surge of interest in the signal-processing community for several reasons. First, the conditions on $\mathcal{G}_1$ and $\mathcal{G}_2$ in \eqref{eq:ADMMstdproblem} (\ie, closed, proper and convex) are mild and \eqref{eq:ADMMstdproblem} therefore encompasses a large number of optimization problems as particular cases. Second, the ADMM recursion \eqref{eq:ADMMstdformA}-\eqref{eq:ADMMstdformC} converges to a solution of \eqref{eq:ADMMstdproblem} under very general conditions, see \cite[section 3.2]{Boyd11}. Third, although ADMM is known to be slow to converge to a solution with high accuracy, it has been shown empirically that ADMM converges to modest accuracy in a few tens of iterations. \\

\section{Algorithm's Details}

\subsection{First Building Block: Computation of   \eqref{eq:grad1}-\eqref{eq:foward1}}\label{sec:app2}

 In this appendix, we complement the exposition done in section \ref{sec:3.1} on the fast  evaluation of   gradient  of cost function $\mathcal{J}(\epsi,\mathbf{d},\mathbf{c})$  given in \eqref{eq:costfunctionun}, particularized to the model parameters  specified in section \ref{sec:4.1}.
First of all, we  expose the particularization of recursions \eqref{eq:grad1}-\eqref{eq:foward1}  to this setting. It is straightforward to see that it  results in the following procedure: \\

\begin{itemize}
\item[\textit{i})] {Compute sequence $\{\xx_t'\}_{t=0}^T$ by  the backward recursion}:\\ 
$\left\{\begin{aligned}
 &\xx_T'=\mathbf{D}\cv',\\
 &\xx_t'=\mathcal{P}(\xx_{t+1}',\mathbf{d}_{t+1}')+\epsi_{t+1}'.\\
  \end{aligned}\right.\\
  $
\item[\textit{ii})] {Compute sequence $\{\zv_t\}_{t=0}^T$ by the forward recursion:}

$\left\{\begin{aligned}
&\zv_0=2 \nabla_{\xx_0}\mathcal{H}^*(\mathcal{H}(\xx_0')-\yy_0),\\
&\zv_{t+1}=  \nabla_{\xx_{t+1}}\mathcal{P}^* (\xx_{t+1}', \dd_{t+1}') \zv_{t}, +2\nabla_{\xx_{t+1}}\mathcal{H}^*(\mathcal{H}(\xx_{t+1}')-\yy_{t+1}).
\end{aligned}\right.$
\item[\textit{iii})]  {Compute the gradients:} \\
$
\left\{\begin{aligned}
&\nabla_{\boldsymbol{\epsilon}_{t}} \mathcal{J}(\epsi',\dd',\cv')= \zv_{t-1}+ 2\alpha_1    \epsi_t',\\  
&\nabla_{\dd_{t}} \mathcal{J}(\epsi',\dd',\cv')=\mathbf{E}_t
  \zv_{t-1} +2\alpha_2\mathbf{W}_t \mathbf{G}\mathbf{G}^* \dd_{t}',\\
&\nabla_{\cv} \mathcal{J}(\epsi',\dd',\cv')=
\mathbf{D}^* \zv_T+2\alpha_3  \cv'
\end{aligned}\right.\\[0.5cm]
$
\end{itemize} 
where  $\mathbf{W}_t \in \Rr^{2n \times 2n}$ and $\mathbf{E}_t  \in \Rr^{2n \times n}$ are respectively  diagonal and block-diagonal matrices which will be defined in the following. 
 We  detail  hereafter the elements of the procedure which  have not been fully described yet. 

We begin by making some comments  on the evaluation of the warping function  $\mathcal{P}(\xx_{t}',\mathbf{d}_{t}')$ and  its Jacobian $\nabla_{\xx_t} \mathcal{P}(\xx_{t}',\mathbf{d}_{t}') $, which  constitute the core of the recursion. 
 We propose to use   the family of bi-dimensional cubic cardinal   splines  $\{\psi_i\}_{i=1}^n$ for the representation \eqref{eq:Pdx}.
In practice, we compute an equivalent representation  based on   the family of bi-dimensional  cubic  B-splines functions $\{\phi_i\}_{i=1}^n$. Indeed,  this  representation presents some computational advantages because of the existence of fast B-splines transforms. The relation between cardinal  cubic splines and cubic  B-splines functions is given in \cite{Unser91}. This reference also provides details on the fast cubic  B-splines transform by recursive filtering. 
Let matrix $\mathbf{C}^*=[\mathbf{c}_1,...,\mathbf{c}_n]^* \in\Rr^{n \times n}$ denote the direct B-spline transform  of a discrete bi-dimensional signal, \ie, the transform computing from a discrete signal $\xx_t$ its representation with spline coefficients $ \mathbf{C}^*\xx_t$.   Rewritten \eqref{eq:Pdx}   with cubic B-spline functions, we get: 
 \begin{align}\label{eq:splineRep}
   \mathcal{P}_s(\xx_{t},{\mathbf{d}_{t}})& = \sum_{i \in \vartheta(\chi(s)+\dd_t({s}))} \mathbf{c}_i^*\xx_t \, \phi_i(\chi(s)+\mathbf{d}_{t}(s)),  
    \end{align}
   where   $\vartheta(\chi(s)+\dd_t({s}))$ denotes a subset of  vector indices corresponding to the neighborhood of the spatial position $\chi(s)$ 
    (which differs from the subset $\mathcal{V}$ previously defined  in   \eqref{eq:Pdx}). To simplify notations, we denote by $\mathcal{I}:\Rr^n \times \Rr^{2n} \to \Rr^n$ the function taking as a first argument spline coefficients   $\mathbf{C}^*\xx_t$ and as a second argument a motion field $\dd_t$, and whose $s$-th component is  given by \eqref{eq:splineRep}. 
   Using this  notation,   \eqref{eq:splineRep} can be rewritten in the  vectorial form
     \begin{align*}
    \mathcal{P}(\xx_{t},{\mathbf{d}_{t}}) & =\mathcal{I}(\mathbf{C}^*\xx_t,\mathbf{d}_{t}).
 \end{align*}
We denote by   $\nabla \mathcal{I}( \mathbf{C}^*\xx_t,\dd_{t})$ the Jacobian of function $\mathcal{I}$ at point $(\mathbf{C}^*\xx_t,\dd_{t})$ with respect to its first argument, \ie, spline coefficients.  Since  function $\mathcal{I}$ is linear with respect to  spline coefficients, the Jacobian is only dependent on the value of its second argument, \ie, $\dd_{t}$. Therefore, we will adopt  the notation $\nabla \mathcal{I}(\dd_{t})$ in the sequel.  

The complexity of evaluating both spline coefficients $\mathbf{C}^*\xx_{t}$ and  the interpolated function $\mathcal{I}$, scales linearly with the image dimension, \ie, $\mathcal{O}(n)$, thanks to the representation separability and to recursive linear filtering \cite{Unser91}. Multiplication with the Jacobian   transpose
$$
\nabla_{\xx_{t}} \mathcal{P}^* (\xx_{t+1} ,  \dd_{t+1}) = \mathbf{C} \nabla \mathcal{I}^*(\mathbf{d}_{t+1}),
 $$
  implies also a linear complexity: first,  matrix $\mathbf{C}$ is symmetric\footnote{Matrix $\mathbf{C}$ is symmetric in the case of periodic boundary conditions  \cite{Unser91}.} 
  so that it is identical to  the direct B-spline transformation  $\mathbf{C}^* $, computed by recursive linear filtering; second, the multiplication of the Jacobian transpose  of function $\mathcal{I}$  with vector $\zv_t$ is equal to
  $$\nabla \mathcal{I}^*(\mathbf{d}_t) \zv_{t}(s)=\sum_{i | s \in \vartheta(\chi(i)+\dd_t({i}))}  \zv_{t} (i) \phi_s(\chi(i)+\dd_t({i})).$$

Concerning the Jacobian  transpose $\nabla_{\xx_t} \mathcal{H}^*$,  it is easy to see that 
this matrix is an up-sampling operation, inserting zeros,  followed by the same  low-pass filtering as  in $\mathcal{H}$. 

We continue by detailing matrices appearing in the last step of the procedure.  First, we note that  matrix $\mathbf{D}^*$ is simply the  direct  Real-valued Fast Curvelet Transform. This transform is, as well as its transpose $\mathbf{D}$, based on fast Fourier transforms, whose complexity scales in $\mathcal{O}(n\log n)$ \cite{Mallat2008Wavelet}. Next,   the two diagonals of the two-block matrix $\mathbf{E}_t$ are the two  $n$-dimensional vectors   $\partial_{s_j} \left(\mathcal{I}(\mathbf{C}^*\xx_{t},\mathbf{d}_{t})\right) $ for $j=1,2$,   where   $s
 _j$ denotes the $j$-th spatial coordinate.  
We approach  these partial derivatives by second-order centered finite differences. Then, the  diagonal of  matrix $\mathbf{W}_t $ is the vector concatening twice the  weight vector $\mathbf{w}_t $, \ie,   $\mathbf{W}_t(s,s)=\mathbf{W}_t(2s,2s)=\mathbf{w}_t(s)$ for $s=1,...,n$. 


To finalize the description of this procedure, it remains to give some  details on matrix $\mathbf{G}$.
Let the elements of vector $\mathbf{G}^*\mathbf{d}_t$ be  first-order forward finite-difference approximations  of the spatial gradients of the two  motion  components, which have been rearranged beforehand on the pixel grid.  
This gradient approximation becomes exact assuming that components of vector $\mathbf{d}_t$ are  coefficients associated to the decomposition of some continuous motion field in a  basis of  interpolating and separable scaling functions (see a proof in \cite{Lemariet92}).   Straightforward calculus then shows  that  elements of vector $\mathbf{G}\mathbf{G}^*\mathbf{d}_t$ are  second-order finite difference approximations  of the Laplacian  of  the two  motion  components, which have been rearranged beforehand on the pixel grid.

\subsection{Second Building Block:  ADMM Solver for  Problems \eqref{eq:Puy1} and \eqref{eq:Puy2}}\label{app:attouch}
In this appendix, we present an ADMM implementation of the two minimization problems \eqref{eq:Puy1} and \eqref{eq:Puy2} appearing in the procedure described in section \ref{sec:3.3} (which also corresponds to Algorithm 4 introduced later on in section \ref{sec:4}).  In the following, iterations  of the 2-step recursion presented in section \ref{sec:3.3} will be indexed by the exponent $^{(\ell)}$, in order to differentiate them from the iterations  related to  ADMM, 
 which will  be indexed by the exponent $^{(k)}$. \\

 We begin by  the analysis of minimization problem \eqref{eq:Puy1}. This problem can  be equivalently reexpressed as:
\begin{align}\nonumber
\argmin_{(\xx,\epsi, \mathbf{c})\in\Cc,(\tilde{\epsi}, \tilde{\mathbf{c}})} &\sum_{t=0}^{T}\| \mathcal{H} (\xx_t )-\yy_t \|^2_2
  +  \sum_{t=1}^{T}\left( \alpha_1 \|  \tilde{\epsi}_t  \|_1 +\gamma\| \tilde \delta_{\epsi_t} \|_1 \right) 
  + \alpha_3\| \tilde{\cv}\|_1 +\gamma\, \|\tilde \delta_{\cv} \|_1\\
 & \textrm{s.t.}\ \left\{ \nonumber
\begin{array}{ll}
\epsi_t=\tilde{\epsi_t},\quad\quad\quad\quad\quad\quad \forall t,\\
 \mathbf{H}^* (\epsi_t-\epsi^{(\ell)}_t)= \tilde \delta_{\epsi_t},\quad \forall t,\\
\mathbf{c}=\tilde{\mathbf{c}},\\
 \cv- \boldsymbol{\cv}^{(\ell)}= \tilde \delta_{\cv},\\
 \end{array}
  \right.
\end{align}
where
\begin{align*}
\Cc  \triangleq \left\{ (\xx,\epsi, \mathbf{c}) \left| 
\begin{aligned}
\xx_t=\mathcal{P}(\xx_{t+1},\dd_{t+1})+ \epsi_{t+1},&\hspace{0.2cm} 0\leq t \leq  T-1 \\
\xx_T=\mathbf{D} \cv&
\end{aligned}
\right.
\right\}.
\end{align*}

Here, we have added four new variables to the problem, $\tilde{\epsi}=(\tilde{\epsi}_1,...,\tilde{\epsi}_T)$, $\tilde{\mathbf{c}}$, $\tilde \delta_{\epsi}=(\tilde\delta_{\epsi_1},...,\tilde\delta_{\epsi_T})$ and $ \tilde \delta_{\cv}$, which are counterbalanced by the inclusion of four new constraints.
We use the formalism exposed in appendix \ref{app:ADMM} with $\zz_1=(\xx,\epsi,  \mathbf{c})$, $\zz_2=(\tilde{\epsi}, \tilde{\mathbf{c}},\tilde  \delta_{\epsi},   \tilde \delta_\cv)$, $\Zc_1=\Cc$ and  $\Zc_2=\Rb^{nT} \times \Rb^q \times \Rb^{nT} \times \Rb^q$ and obtain the following ADMM recursions: 
\begin{align}
&(\xx^{(k+1)}, \epsi^{(k+1)}, \mathbf{c}^{(k+1)}) = \argmin_{(\xx,\epsi,\mathbf{c}) \in \Omega}\mathcal{L}^{(k)} (\xx,\epsi,\mathbf{c})+ \frac{\rho}{2}\sum_{t=1}^T\|\mathbf{H}^* (\epsi_t-\epsi^{(\ell)}_t)  -  \tilde \delta_{\epsi_t}^{(k)}+  \mathbf{u}_{ \delta_{\epsi_t}}^{(k)} \|^2_2 \nonumber \\
&\quad\quad\quad\quad\quad\quad\quad\quad\quad\quad\quad\quad\quad\quad 
 + \frac{\rho}{2}\|  \cv- \boldsymbol{\cv}^{(\ell)}-  \tilde \delta_\cv^{(k)}+  \mathbf{u}_{{\delta}_\cv}^{(k)} \|^2_2, \label{eq:step1_ADMM_Attouch1} \\
 & 
 \left\{
\begin{array}{ll}
\tilde{\epsi}^{(k+1)}_t =& \argmin_{\tilde{\epsi}_t}\  \|  \tilde{\epsi}_t  \|_1 + \frac{\rho_1}{2 \alpha_1} \| \epsi^{(k+1)}_t- \tilde{\epsi}_t   + \mathbf{u}_{\epsi_t}^{(k)} \|^2_2, \\
\tilde{\mathbf{c}}^{(k+1)}=& \argmin_{\tilde{\mathbf{c}}}\quad \| \tilde{\cv}\|_1+ \frac{\rho_3}{2 \alpha_3} \|  \mathbf{c}^{(k+1)} - \tilde{\mathbf{c}} + \mathbf{u}_{\mathbf{c}}^{(k)} \|^2_2,\\
\tilde \delta_{\epsi_t}^{(k+1)} =& \argmin_{\tilde \delta_{\epsi_t} }\  \|  \tilde \delta_{\epsi_t}  \|_1 + \frac{\rho}{2 \gamma} \| \mathbf{H}^* (\epsi^{(k+1)}_t-\epsi^{(\ell)}_t )-\tilde \delta_{\epsi_t}   + \mathbf{u}_{\delta_{\epsi_t}}^{(k)} \|^2_2,  \\
\tilde \delta_\cv^{(k+1)}=& \argmin_{\tilde \delta_\cv}\quad \| \tilde \delta_\cv\|_1+ \frac{\rho}{2 \gamma} \|  \cv^{(k+1)}- \boldsymbol{\cv}^{(\ell)} - \tilde \delta_\cv + \mathbf{u}_{{\delta}_\cv}^{(k)} \|^2_2,
\end{array}\label{eq:step2_ADMM_Attouch1}
 \right.\\
 &
 \left\{
 \begin{array}{ll}
 \mathbf{u}_{\epsi_t}^{(k+1)} =& \mathbf{u}_{\epsi_t}^{(k)} + \epsi^{(k+1)}_t - \tilde{\epsi}^{(k+1)}_t,\\
\mathbf{u}_{\mathbf{c}}^{(k+1)}=& \mathbf{u}_{\mathbf{c}}^{(k)} + \mathbf{c}^{(k+1)}- \tilde{\mathbf{c}}^{(k+1)}, \\
\mathbf{u}_{\delta_{\epsi_t}}^{(k+1)} =&\mathbf{u}_{\delta_{\epsi_t}}^{(k)} + \mathbf{H}^*(\epsi^{(k+1)}_t -\epsi^{(\ell)}_t) - \tilde \delta_{\epsi_t}^{(k+1)},\\
\mathbf{u}_{{\delta}_\cv}^{(k+1)}=&\mathbf{u}_{{\delta}_\cv}^{(k)} + \mathbf{c}^{(k+1)}- \boldsymbol{\cv}^{(\ell)}-\tilde \delta_\cv^{(k+1)},
\end{array}\label{eq:step3_ADMM_Attouch1}
 \right.
\end{align}
where  $\mathcal{L}^{(k)}$ is defined in \eqref{eq:Lk}. 
Equations \eqref{eq:step1_ADMM_Attouch1}, \eqref{eq:step2_ADMM_Attouch1} and \eqref{eq:step3_ADMM_Attouch1} correspond respectively to expressions \eqref{eq:ADMMstdformA}, \eqref{eq:ADMMstdformB} and \eqref{eq:ADMMstdformC} in appendix \ref{app:ADMM}. 
We comment on the two first  steps of the ADMM algorithm, the last one being trivial. 
 First, as already mentioned, problem \eqref{eq:step1_ADMM_Attouch1} has the same structural form as the problem addressed in section \ref{sec:3.1}. We thus  apply the methodology described in section \ref{sec:3.1} to solve this problem via a gradient descent algorithm.
      The core of this methodology is  the computation of the gradient of the cost function with respect to $\mathbf{c}$ and $\epsi$. 
The gradient efficient evaluation relies on a  backward-forward recursion possessing the  structural form of the first building block constituted by equations  \eqref{eq:grad1}-\eqref{eq:foward1}. Some details of the implementation on  \eqref{eq:grad1}-\eqref{eq:foward1} are provided in appendix \ref{sec:app2}, for the particular case of the model parameters given in section \ref{sec:4.1}.

 We remark that the complexity associated to the evaluation of the gradient scales as $\mathcal{O}(nT+q)$. 
Second, the optimization problems specified in \eqref{eq:step2_ADMM_Attouch1} all have  simple analytical solutions based on soft-thresholding operators \eqref{eq:softthresh}. 
 We immediately remark that the two first updates in  \eqref{eq:step2_ADMM_Attouch1}  are identical to the ADMM steps  \eqref{eq:step2_ADMM}  used to treat  the convex case in section \ref{sec:3.2}. Moreover, the solutions to the last two problems  in \eqref{eq:step2_ADMM_Attouch1} are given by 
\begin{align}
\left\{
\begin{array}{ll}
\tilde \delta_{\epsi_t}^{(k+1)}(i)&=\mathrm{soft}_{\frac{\gamma}{\rho}} \left( \mathbf{h}_i^*(\epsi^{(k+1)}_t-\epsi^{(\ell)}_t) + \mathbf{u}_{\delta_{\epsi_t}}^{(k)}(i)\right)\\
\tilde \delta_\cv^{(k+1)}(i)&=\mathrm{soft}_{\frac{\gamma}{\rho}} \left( \cv^{(k+1)}(i)- \boldsymbol{\cv}^{(\ell)}(i) +  \mathbf{u}_{{\delta}_\cv}^{(k)}(i) \right)
\end{array}
\right. \qquad\forall i,
\end{align}
where $\mathbf{h}_i$ is the $i$th column of $\mathbf{H}$ and ``$\mathrm{soft}$'' denotes the soft-threshloding operator defined in \eqref{eq:softthresh}. 
 
We continue  with the analysis of  minimization  problem \eqref{eq:Puy2}. We first remark that   we can  apply the methodology described in section \ref{sec:3.1} to compute  the gradient $\nabla_{\dd_t} \mathcal{B}(\dd^{(\ell)})$ required to build the quadratic approximation \eqref{eq:defquadapprox}. 
 Once this quadratic approximation has been obtained, the task now is to solve minimization problem \eqref{eq:Puy2}.
 We can notice that this problem does unfortunately not possess an explicit solution. 
 To circumvent this issue, we  use an ADMM strategy, 
 as detailed below. Problem  \eqref{eq:Puy2} is  reexpressed as:
\begin{align}\nonumber
&\argmin_{\dd,(\tilde{\dd}_1,...,\tilde{\dd}_T)} \,  \mathcal{B}(\mathbf{d},\mathbf{d}^{(\ell)})
   + \alpha_2 \sum_{t=1}^{T}  \mathcal{R}(\tilde\dd_t  )\\
 & \textrm{s.t.}\quad 
\mathbf{G}^*\dd_t=\tilde{\dd_t},\quad\quad\forall t.
\end{align}
Here, we have added the new variables $\tilde{\dd}_t$'s to the problem  which are counterbalanced by the inclusion of  new constraints.
We use the formalism exposed in appendix \ref{app:ADMM} with $\zz_1=(\dd_1,...,\dd_T)$, $\zz_2=(\tilde{\dd}_1,...,\tilde{\dd}_T)$,  $\Zc_1=\Rr^{2nT}$  and  $\Zc_2=\Rb^{hT}$ and obtain the following ADMM recursions:
\begin{align}
&\dd^{(k+1)} = \argmin_{\dd} \mathcal{B}(\mathbf{d},\mathbf{d}^{(\ell)})  + \frac{\rho_2}{2} \sum_{t=1}^{T}\|  \mathbf{G}^*\dd_t-  \tilde \dd_t^{(k)}+  \mathbf{u}_{\dd_t}^{(k)} \|^2_2,\label{eq:step1_ADMM_Attouch2} \\
&\tilde\dd_t ^{(k+1)}= \argmin_{\tilde\dd_t }\quad   \mathcal{R}(\tilde\dd_t  )+ \frac{\rho_2}{2 \alpha_2}\|  \mathbf{G}^*\dd_t^{(k+1)}-  \tilde \dd_t+  \mathbf{u}_{\dd_t}^{(k)} \|^2_2, \label{eq:step2_ADMM_Attouch2}\\
 &
\mathbf{u}_{\dd_t}^{(k+1)}=\mathbf{u}_{\dd_t}^{(k)} +\mathbf{G}^*\dd_t^{(k+1)}-\tilde\dd_t ^{(k+1)}.\label{eq:step3_ADMM_Attouch2}
\end{align}
Equations \eqref{eq:step1_ADMM_Attouch2}, \eqref{eq:step2_ADMM_Attouch2} and \eqref{eq:step3_ADMM_Attouch2} correspond respectively to expressions \eqref{eq:ADMMstdformA}, \eqref{eq:ADMMstdformB} and \eqref{eq:ADMMstdformC} in appendix \ref{app:ADMM}. 

We comment now on the resolution of \eqref{eq:step1_ADMM_Attouch2} and \eqref{eq:step2_ADMM_Attouch2}. First, the unconstrained differentiable  problem \eqref{eq:step1_ADMM_Attouch2} can be easily solved via a gradient descent algorithm. The gradient of the cost function in \eqref{eq:step1_ADMM_Attouch2} with respect to $\mathbf{d}_t$ can be expressed as
\begin{align}
\nabla_{\dd_t} \mathcal{B}(\dd^{(\ell)})+ \alpha^{(\ell)}(\dd_t-\dd_t^{(\ell)})+\rho_2 \mathbf{G}(\mathbf{G}^*\dd_t- \tilde \dd_t^{(k)}+  \mathbf{u}_{\dd_t}^{(k)}).\label{eq:gradADMMAttouch}
\end{align}
As mentioned previously,  $\nabla_{\dd_t} \mathcal{B}(\dd^{(\ell)})$ is simple to evaluate via the recursions described in section \ref{sec:3.1}; moreover, the multiplications by $\mathbf{G}$ and $\mathbf{G^*}$ appearing in the last term of \eqref{eq:gradADMMAttouch} can be done efficiently for the particular choice of $\mathbf{G}$ considered in this paper (see section \ref{sec:app2} for details on this topic).

Second,  the solution of  problem \eqref{eq:step2_ADMM_Attouch2}  is closed-form (see \eg, \cite[section 6.5.2]{OPT-003}). It is given for any $  j \in \mathcal{S}_i$ with $1\le i\le n$ by  
 \begin{equation*}
\tilde\dd_t^{(k+1)}(j)=\left\{\begin{aligned}
&0\hspace{0.3cm}\textrm{if }\tau_i\le \alpha_2 \mathbf{w}(i)/\rho_2, \\
&\frac{\left(\tau_i- {\alpha_2}  \mathbf{w}(i)/{\rho_2}\right)}{\tau_i}(\mathbf{g}_j^* \dd^{(k+1)}_{t} + \mathbf{u}_{\dd_t}^{(k)}(j))\hspace{0.3cm}\textrm{otherwise,}\\
\end{aligned}\right. 
\end{equation*}
where the scalar $\tau_i$ is given by $$\tau_i=\sqrt{ \sum_{ j \in \mathcal{S}_i}\left(\mathbf{g}_j^* \dd^{(k+1)}_{t} + \mathbf{u}_{\dd_t}^{(k)}(j)\right)^2}.$$

%
%
%

 \bibliographystyle{spmpsci}      
\bibliography{ref,group-15302,cherzet}

\newpage
   
\begin{figure}[!ht]

\begin{center}
\begin{tabular}{cccc}
\includegraphics[width=0.5\textwidth]{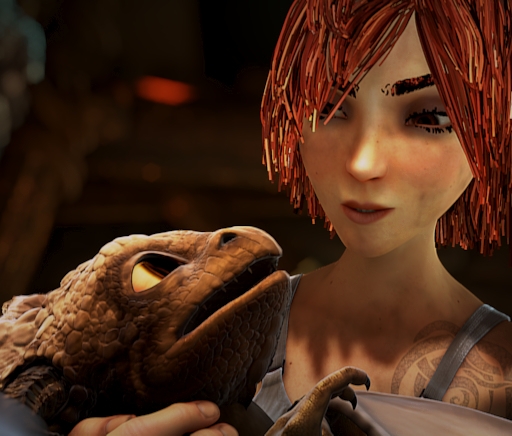}&
\includegraphics[width=0.5\textwidth]{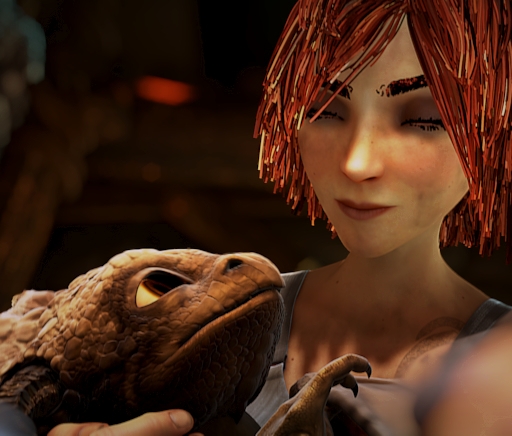}\\
\end{tabular}
	\caption{{\footnotesize    \textbf{Data set \#1}:  first and last frames of the {``\textit{bandage}''} sequence.}  \label{figNew0}}\vspace{-0.cm}
	\end{center}
\end{figure}

\noindent

\begin{figure}[!ht]
\begin{center}
\begin{tabular}{cccc}
\includegraphics[width=0.25\textwidth]{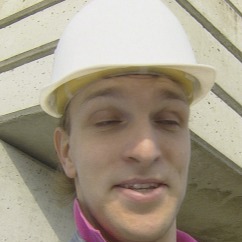}&
\includegraphics[width=0.25\textwidth]{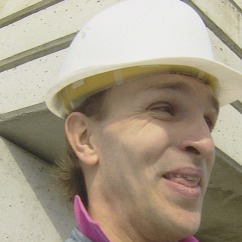}&
\end{tabular}
	\caption{{\footnotesize    \textbf{Data set  \#2 }:   first and last  frames of  the  {``\textit{foreman}''}  sequence.}  \label{figNew1}}\vspace{-0.cm}
	\end{center}
\end{figure}
\noindent

\begin{figure}[!ht]
\begin{center}
\begin{tabular}{cccc}
\includegraphics[width=0.275\textwidth]{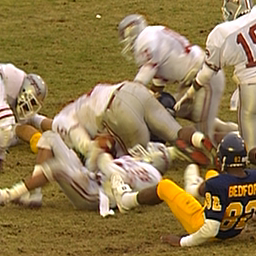}&
\includegraphics[width=0.275\textwidth]{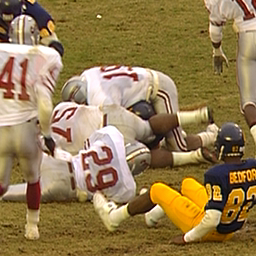}
\end{tabular}
	\caption{{\footnotesize    \textbf{Data set  \#3}:   first and last  frames of  the {``\textit{football}''} sequences.}  \label{figNew2}}\vspace{-0.cm}
	\end{center}
\end{figure}
\noindent

\begin{figure}	[!ht]
\begin{center}
\begin{tabular}{ccc}
\includegraphics[width=0.30\textwidth]{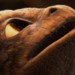} &\includegraphics[width=0.30\textwidth]{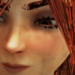}&\includegraphics[width=0.30\textwidth]{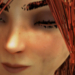}\\
\includegraphics[width=0.30\textwidth]{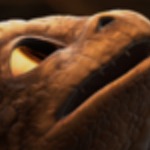}& \includegraphics[width=0.30\textwidth]{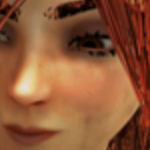}&\includegraphics[width=0.30\textwidth]{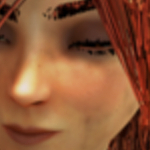}  \\
\includegraphics[width=0.30\textwidth]{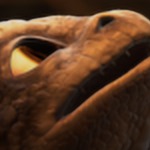} & \includegraphics[width=0.30\textwidth]{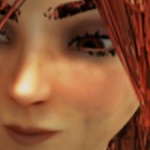}& \includegraphics[width=0.30\textwidth]{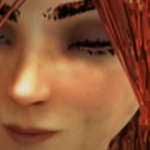} \\
\end{tabular}
	\caption{{\footnotesize  \textbf{``Single-image'' SR estimates  for  data set \#1.}  Details of the SR images obtained with nearest neighbor strategy ($1$-st row),  Lanczos interpolation ($2$-nd row) and  the learning algorithm proposed in \cite{Peleg14}    ($3$-rd row).  \label{figNew3}}}\vspace{-0.cm}
		\end{center}
\end{figure}

\begin{figure}	[!ht]
\begin{center}
\begin{tabular}{ccc}
 \includegraphics[width=0.30\textwidth]{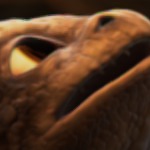}& \includegraphics[width=0.30\textwidth]{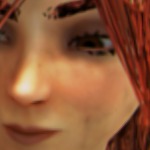}& \includegraphics[width=0.30\textwidth]{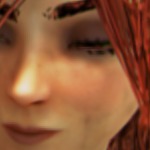}   \\
\includegraphics[width=0.30\textwidth]{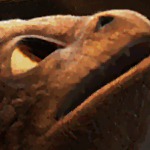}&\includegraphics[width=0.30\textwidth]{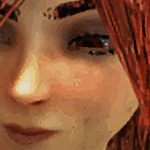}& \includegraphics[width=0.30\textwidth]{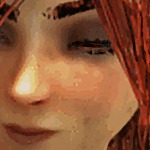}  \\
\includegraphics[width=0.30\textwidth]{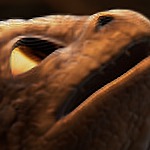} &\includegraphics[width=0.30\textwidth]{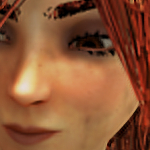}&\includegraphics[width=0.30\textwidth]{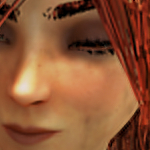}\\
\hline \\
\includegraphics[width=0.30\textwidth]{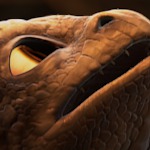} &\includegraphics[width=0.30\textwidth]{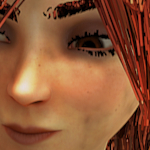}&\includegraphics[width=0.30\textwidth]{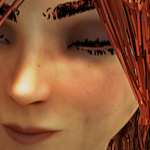}
\end{tabular}
	\caption{{\footnotesize   \textbf{Multi-frame  and sequential SR estimates for  data set \#1.}  Details of the SR images obtained with the multi-frames algorithms  of \cite{Takeda09,Milanfar10}  ($1$-st row) or   \cite{Mitzel09}  ($2$-nd row), and with the proposed sequential algorithm   ($3$-rd row)  in comparison to ground truth  ($4$-th row).   { Initialization of our algorithm relies on the optic-flow method  \cite{Xu12}. }  \label{figNew4}}}\vspace{-0.cm}
		\end{center}
\end{figure}

\begin{figure}	[!ht]
\begin{center}
\hspace{-0cm}\begin{tabular}{cccc}
\includegraphics[width=0.25\textwidth]{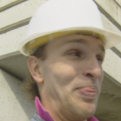} &\includegraphics[width=0.25\textwidth]{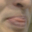}&\includegraphics[width=0.25\textwidth]{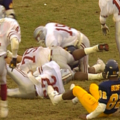} &\includegraphics[width=0.25\textwidth]{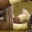}\\
\includegraphics[width=0.25\textwidth]{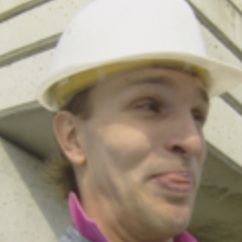} &\includegraphics[width=0.25\textwidth]{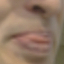}&\includegraphics[width=0.25\textwidth]{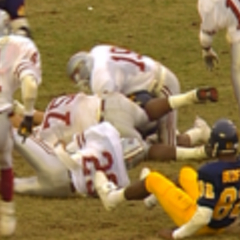} &\includegraphics[width=0.25\textwidth]{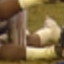}\\
\includegraphics[width=0.25\textwidth]{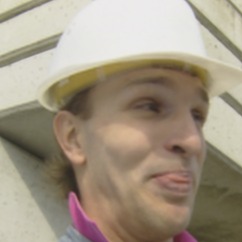} & \includegraphics[width=0.25\textwidth]{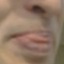}&\includegraphics[width=0.25\textwidth]{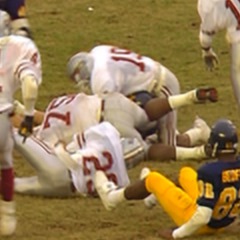} & \includegraphics[width=0.25\textwidth]{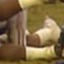}\\
\end{tabular}
	\caption{{\footnotesize  \textbf{``Single-image'' SR estimates  for  data set \#2 and \#3.}  SR images and details obtained with nearest neighbor strategy ($1$-st row),  Lanczos interpolation ($2$-nd row) and  the learning algorithm of \cite{Peleg14}    ($3$-rd row).  \label{figNew6}}}\vspace{-0.cm}
		\end{center}
\end{figure}

\begin{figure}	[!ht]
\begin{center}
\begin{tabular}{cccc}
\includegraphics[width=0.25\textwidth]{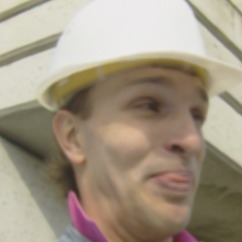} &\includegraphics[width=0.25\textwidth]{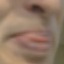} &\includegraphics[width=0.25\textwidth]{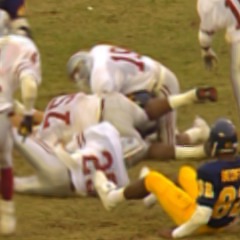}&\includegraphics[width=0.25\textwidth]{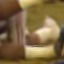}\\
\includegraphics[width=0.25\textwidth]{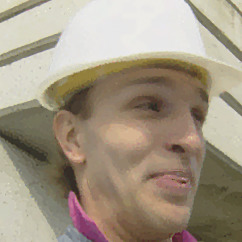} &\includegraphics[width=0.25\textwidth]{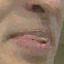} &\includegraphics[width=0.25\textwidth]{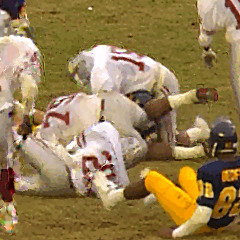}&\includegraphics[width=0.25\textwidth]{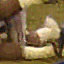}\\
\includegraphics[width=0.25\textwidth]{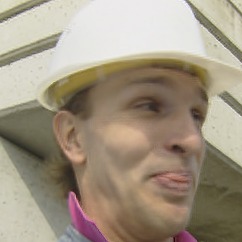}&\includegraphics[width=0.25\textwidth]{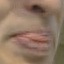}& \includegraphics[width=0.25\textwidth]{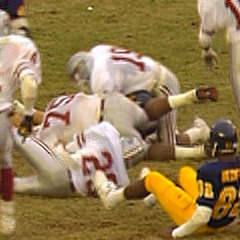}& \includegraphics[width=0.25\textwidth]{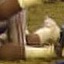}\\
\includegraphics[width=0.25\textwidth]{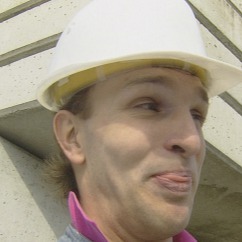} &\includegraphics[width=0.25\textwidth]{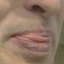} & \includegraphics[width=0.25\textwidth]{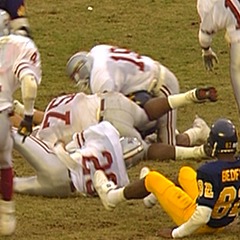}& \includegraphics[width=0.25\textwidth]{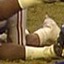}
\end{tabular}
	\caption{{\footnotesize  \textbf{Multi-frame  or sequential SR estimates for  data sets \#2 and \#3.}  SR images  and details  obtained with the multi-frame algorithms  of \cite{Takeda09,Milanfar10}  ($1$-st row) or   \cite{Mitzel09}  ($2$-nd row), and with the proposed sequential algorithm  ($3$-rd row)  in comparison to ground truth  ($4$-th row).  { Initialization of our algorithm relies on the optic-flow method  \cite{Xu12}. 
	}\label{figNew7}}}\vspace{-0.cm}
		\end{center}
\end{figure}

\begin{figure}	[!ht]
\begin{center}
\begin{tabular}{cc}
\includegraphics[width=0.5\textwidth]{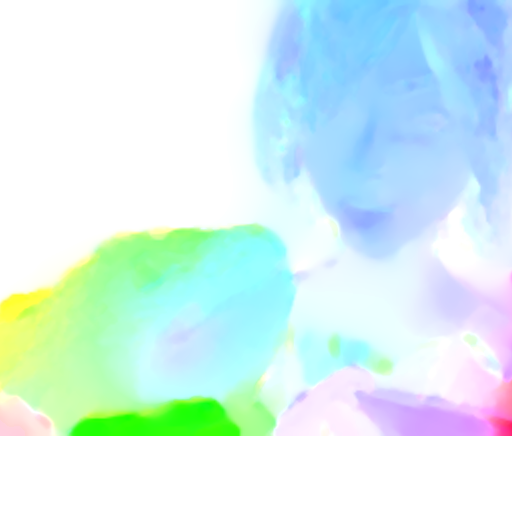}&\includegraphics[width=0.5\textwidth]{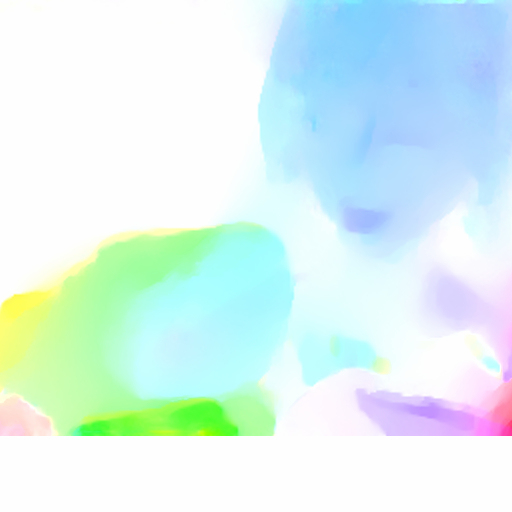}\\
\includegraphics[width=0.5\textwidth]{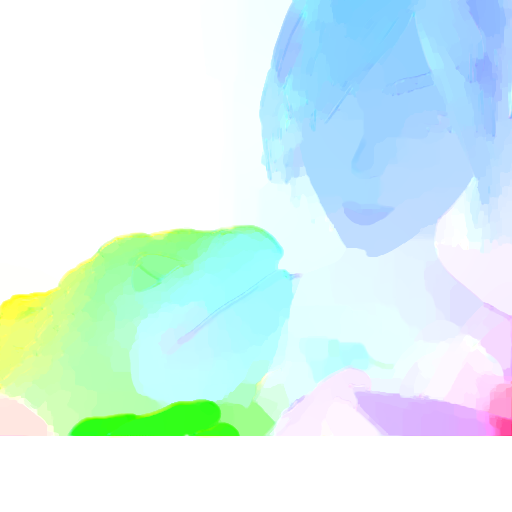}&\includegraphics[width=0.5\textwidth]{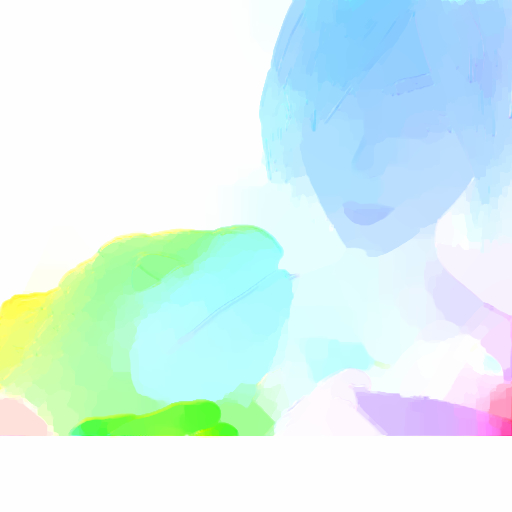}\\
\includegraphics[width=0.5\textwidth]{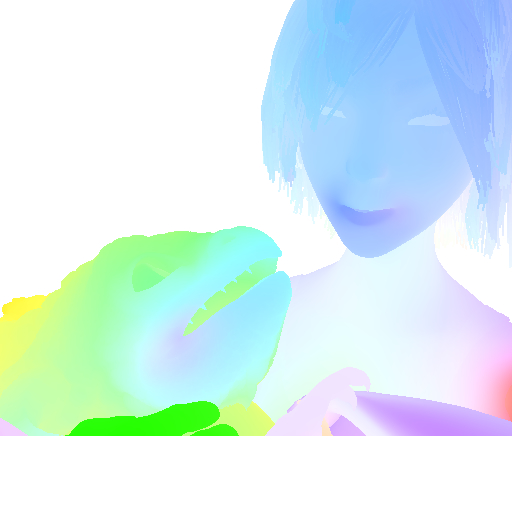}\vspace{-6.25cm}\\
\hspace{-4cm}\includegraphics[width=0.1\textwidth]{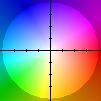}\vspace{+3.75cm}
\end{tabular}
	\caption{{\footnotesize  {  \textbf{Optic-flow SR estimates.} Motion field estimated from low-resolved images  of data set \#1 at initial time. \textbf{Top:}  estimates for state-of-the-art algorithms  \cite{Xu12}  (left) and \cite{Zach07}  (right). \textbf{Middle:}  estimates with the proposed SR algorithm   initialized with  \cite{Xu12} (left) or \cite{Zach07} (right). \textbf{Bottom:}  ground truth and associated colormap.   \label{figNew5}}}}\vspace{-0.cm}
		\end{center}
\end{figure}

\begin{figure}	[!ht]
\begin{center}
\begin{tabular}{ccc}
\includegraphics[width=0.30\textwidth]{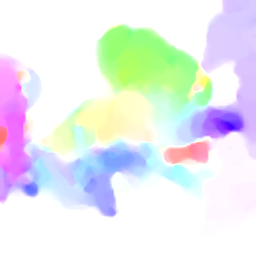} & \includegraphics[width=0.30\textwidth]{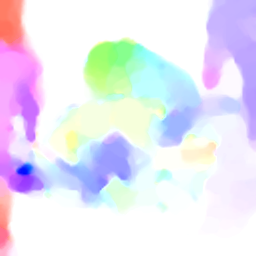}& \includegraphics[width=0.30\textwidth]{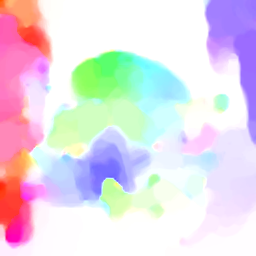} \\
\includegraphics[width=0.30\textwidth]{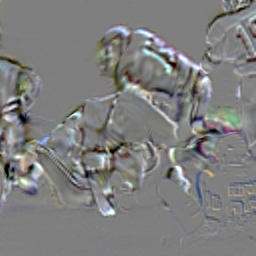} & \includegraphics[width=0.30\textwidth]{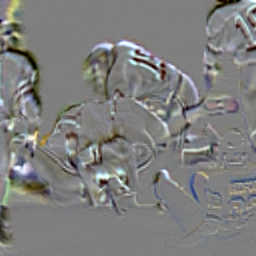}& \includegraphics[width=0.30\textwidth]{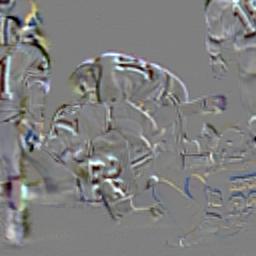} \\
\includegraphics[width=0.30\textwidth]{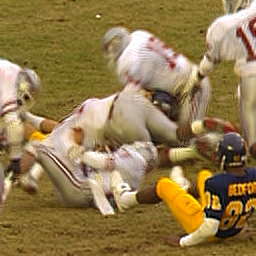}& \includegraphics[width=0.30\textwidth]{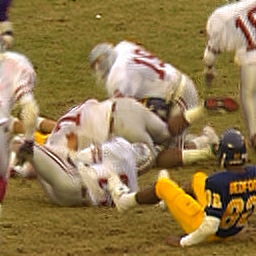}&\includegraphics[width=0.30\textwidth]{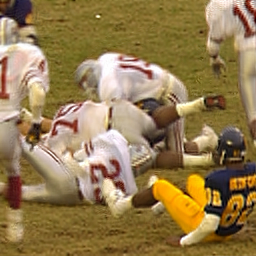}  \\
PSNR=26.989&PSNR=26.743&PSNR=26.560\\
\hline\\
\includegraphics[width=0.30\textwidth]{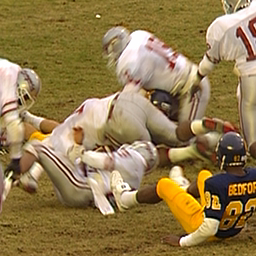} &\includegraphics[width=0.30\textwidth]{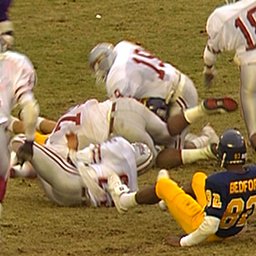}&\includegraphics[width=0.30\textwidth]{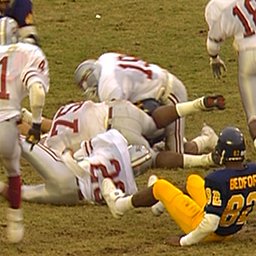}\\
\end{tabular}
\end{center}
\caption{{\footnotesize  Reconstruction of three super-resolved images $\mathbf{x}_t$ ($3$-rd row), optic-flow fields $\mathbf{d}_t$ ($1$-st row) and warping errors $\boldsymbol{\epsilon}_t$ ($2$-nd row)  for  data set \#3   corresponding to $t=3$ (left) $t=5$ (middle) and $t=7$ (right). True images  ($4$-th row) and associated PSNR (computed without quantification of the estimates and including the image borders) are displayed below.     { Initialization of our algorithm relies on the optic-flow method  \cite{Xu12}.} \label{figNew8}} \label{fig:example}}
\end{figure}

\end{document}